\documentclass[runningheads]{llncs}

\usepackage{eccv}

\usepackage{eccvabbrv}

\usepackage{caption}
\usepackage{subcaption}

\usepackage{algorithmic}
\usepackage{algorithm}
\usepackage{amssymb}
\usepackage{threeparttable}
\usepackage{comment}
\usepackage{multirow}

\usepackage{graphicx}
\usepackage{booktabs}

\usepackage[accsupp]{axessibility}  % Improves PDF readability for those with disabilities.

\usepackage{orcidlink}

\usepackage{hyperref}

\begin{document}

\title{View-Consistent Hierarchical 3D Segmentation Using Ultrametric Feature Fields}

% % TODO REVIEW: If the paper title is too long for the running head, you can set
% % an abbreviated paper title here. If not, comment out.
\titlerunning{View-Consistent Hierarchical 3D Segmentation}

% % TODO FINAL: Replace with your author list. 
% % Include the authors' OCRID for the camera-ready version, if at all possible.
\author{Haodi He\orcidlink{0009-0001-5534-7795} \and
Colton Stearns\orcidlink{0000-0002-3297-2870} \and
Adam W. Harley\orcidlink{0000-0002-9851-4645} \and
Leonidas J. Guibas\orcidlink{0000-0002-8315-4886}}

% % TODO FINAL: Replace with an abbreviated list of authors.
\authorrunning{He et al.}


% % TODO FINAL: Replace with your institution list.
\institute{Stanford University}

\maketitle

\begin{abstract}
Large-scale vision foundation models such as Segment Anything (SAM) demonstrate impressive performance in zero-shot image segmentation at multiple levels of granularity. 
However, these zero-shot predictions are rarely 3D-consistent. As the camera viewpoint changes in a scene, so do the segmentation predictions, as well as the characterizations of ``coarse" or ``fine" granularity. 
In this work, we address the challenging task of lifting multi-granular and view-inconsistent image segmentations into a hierarchical and 3D-consistent representation. 
We learn a novel feature field within a Neural Radiance Field (NeRF) representing a 3D scene, whose segmentation structure can be revealed at different scales by simply using different thresholds on feature distance.
Our key idea is to learn an ultrametric feature space, which unlike a Euclidean space, exhibits transitivity in distance-based grouping, naturally leading to a hierarchical clustering.
Put together, our method takes view-inconsistent multi-granularity 2D segmentations as input and produces a hierarchy of 3D-consistent segmentations as output. 
We evaluate our method and several baselines on synthetic datasets with multi-view images and multi-granular segmentation, showcasing improved accuracy and viewpoint-consistency.
We additionally provide qualitative examples of our model's 3D hierarchical segmentations in real world scenes.\footnote{The code and dataset are available at: \url{https://github.com/hardyho/ultrametric_feature_fields}}

\end{abstract}

\section{Introduction}
\label{sec:intro}

Different applications often need different semantic understandings of a scene. This fact necessitates that segmentation methods offer a diverse set of predictions that span different modalities, showcase multiple levels of granularity, and offer hiearchical relationships. 
With the advent of the ``Segment Anything Model'' (SAM)~\cite{kirillov2023segment}, reliable multi-granular \textit{single-view} segmentation might be described as accomplished. 
Yet, in a multi-view or moving-camera system, multi-granular segmentation of each image can produce an overwhelming total number of segments, many of which disagree with each other and most of which are not useful for the downstream application of interest.
In this work, we attempt to distill these thousands of 2D segmentation options (which may be conflicting) into an organized 3D segmentation which is \textit{view-consistent} and \textit{hierarchical}.

\begin{figure}[t]
    \centering
    \includegraphics[width=0.62\linewidth]{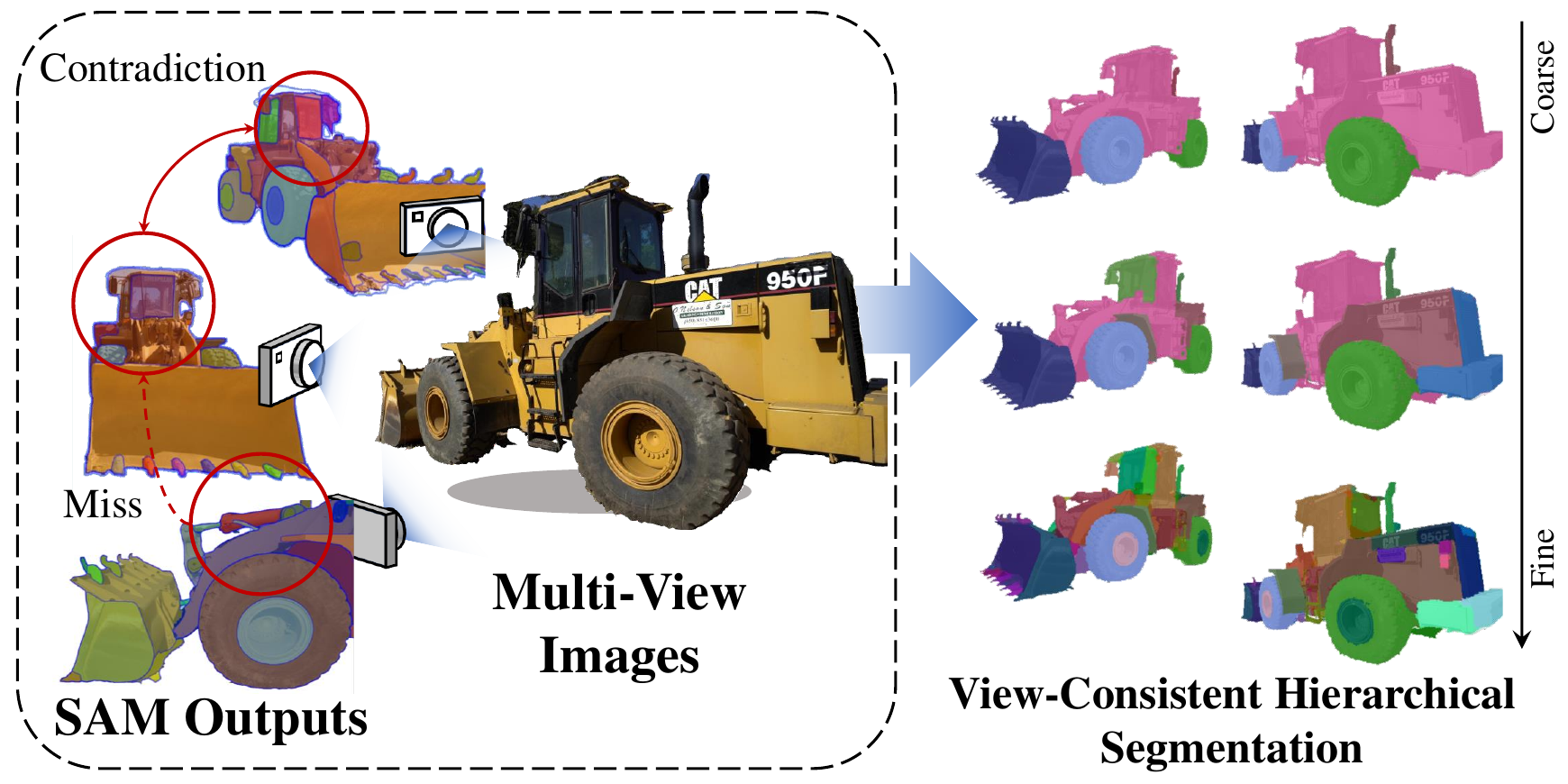}
        
        \caption{Our method takes as input multi-view posed images, paired with segmentation masks from the recent ``Segment Anything Model'' (SAM), and merges these into a coherent 3D representation where segmentation is view-consistent and hierarchical.}
    \label{fig:teaser}
    % \vspace{-3mm}
\end{figure}

As demonstrated in many recent works, \textit{view-consistency} can be achieved by distilling the segmentation information into a 3D implicit field, such as a Neural Radiance Field (NeRF)~\cite{mildenhall2020nerf}.
The first works in this area simply optimized a semantic labelling branch alongside the color branch of the NeRF~\cite{zhi2021inplace,liu2023instance,KunduCVPR2022PNF}. This delivers view-consistent semantic segmentation, but does not generalize to unseen categories. 
More 
recent methods have proposed to distill generic image features into the implicit field~\cite{kobayashi2022distilledfeaturefields,kerr2023lerf}, allowing these features to be queried in 2D or 3D for segmentation or other tasks.  
However, these methods often struggle to capture precise segmentation boundaries, perhaps because language-aligned features are region-based and not pixel-based.
Furthermore, most of these works do not directly account for multiple levels of granularity, and those that do~\cite{kerr2023lerf} must re-render the feature field for every scale of interest. Nonetheless, the expressiveness and wide adoption of NeRFs makes them a natural choice as underlying 3D representation, and we adopt this same choice here. 

Unlike prior work, we explicitly aim for our scene segmentation to be \textit{hierarchical}. 
This means that the scene segmentation has a tree structure, where the root group is the full scene, and any group can be recursively divided into smaller groups, all the way down to the point level. 
A group is defined as a spatially connected neighborhood where all pairwise feature distances are within some threshold. 
At first glance, this is a familiar contrastive learning problem: within-segment feature pairs should have small distances, and cross-segment feature pairs should have large distances, so that thresholding yields groups that follow segmentation boundaries. However, we demonstrate this typical setup is not sufficient to create a consistent hierarchy, and we find it is crucial to use \textit{ultrametric} distances, rather than standard Euclidean distances, in the contrastive loss. 
In an ultrametric space, for any three points $x$, $y$, and $z$, 
distances satisfy a condition stronger than the standard triangle inequality $d(x,z) \le d(x,y)+d(y,z)$, namely that $d(x,z) \le \max\{d(x,y),d(y,z)\}$. 
Ultrametric spaces are ideally suited for hierarchical clustering because distance-based groupings are transitive: if $d(x,y) \le \epsilon$ and $d(y,z) \le \epsilon$, then it follows that $d(x,z) \le \epsilon$. In other words, if we group $x$ and $y$ together and $y$ and $z$ together, then we automatically also group $x$ and $z$ together.
Thus, as $\epsilon$ grows, smaller clusters are naturally merged into larger clusters, automatically giving rise to a hierarchy.

We optimize our 3D ultrametric feature fields by rendering them to 2D, and using SAM as a noisy supervision signal for grouping. 
After optimization, we produce view-consistent hierarchical segmentations at arbitrary levels of granularity, by simply specifying a threshold $\epsilon$, and running a Watershed transform to retrieve all groups that exist under this threshold. 

We evaluate our approach on models from the PartNet dataset, showing that our method recovers a view-consistent hierarchy of segmentations that captures the natural part decomposition of a 3D object. Furthermore, we introduce a synthetic dataset with hierarchical segmentation annotations based on the NeRF Blender Dataset. Unlike PartNet, our proposed dataset offers hierarchical decomposition of more complex scenes. 
In all evaluations, we compare our method to a set of competitive baselines, measuring IoU accuracy and 3D consistency, and we demonstrate that our method outperforms existing open-vocabulary 3D segmentation methods, such as DFF~\cite{kobayashi2022distilledfeaturefields}, LeRF~\cite{kerr2023lerf}, and SAM-3D~\cite{cen2023segment}. Additionally, we introduce a metric for measuring the quality of a hierarchy% (i.e., a soft measure of hierarchical-ness)
, and demonstrate that the segmentations in prior work lack hierarchical structure, while our output is hierarchical by construction.

In sum, our key contribution is a novel formulation for 3D scene segmentation: \textit{ultrametric feature fields}. 
Using this formulation, we are able to distill view-inconsistent 2D masks into a 3D representation which is not only view-consistent but also hierarchical, allowing arbitrary-granularity segmentation at test time. 
We also contribute a new synthetic dataset, and propose new evaluation metrics, to quantify our progress and facilitate future work in this direction. Finally we provide qualitative examples of our model's 3D hierarchical segmentations in real world scenes.

\begin{figure*}[t]
    \centering
    \includegraphics[width=\textwidth]{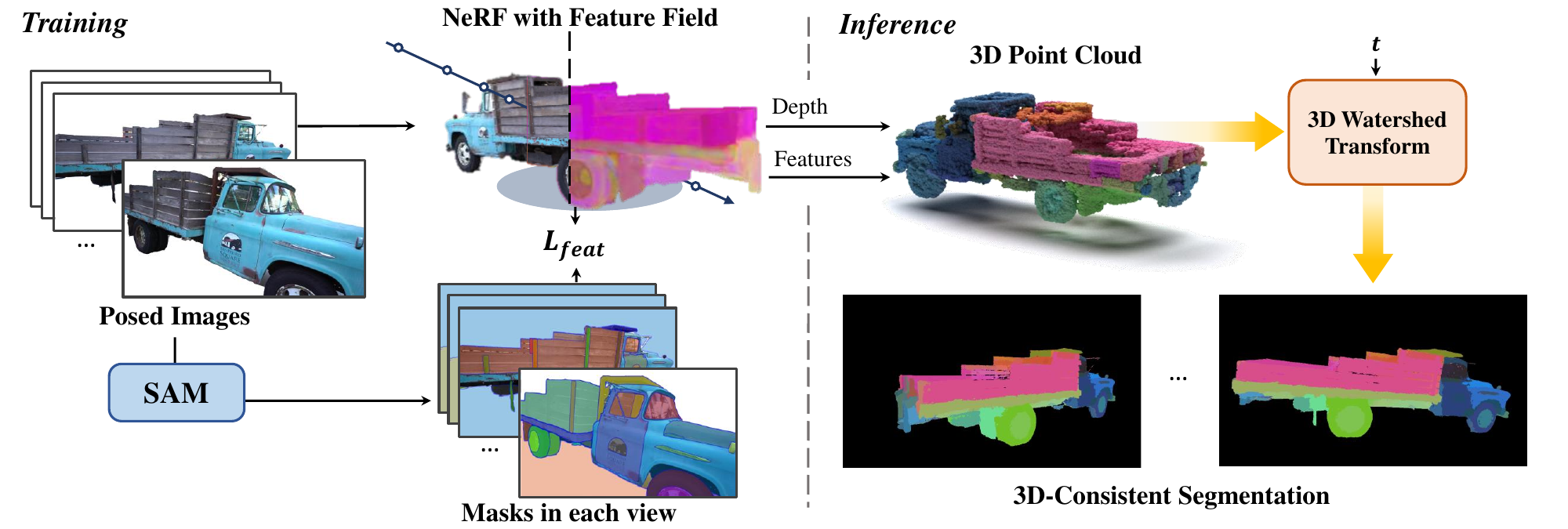}
        
        \caption{\textbf{Method Overview}: We train a NeRF with an ultrametric feature field using images and view-inconsistent segmentation masks from SAM~\cite{kirillov2023segment}. After training, we use the depth estimation and feature maps from training views to construct a 3D point cloud. At inference, for a specified threshold $t$ representing the granularity level, we apply a 3D watershed transform to segment the 3D point cloud. Then, we can query the point clouds in novel views and obtain view-consistent segmentation results. }
    \label{fig:pipeline}
    % \vspace{-3mm}
\end{figure*}

\section{Related Work}
\label{sec:related}

\subsubsection{Neural Radiance and Feature Fields}
Neural Radiance Fields (NeRFs)~\cite{mildenhall2020nerf} are a popular representation for novel view synthesis and 3D reconstruction, and in the last few years there has been an explosion of research surrounding NeRFs.
To name a few research directions, NeRF has been extended to improve rendering quality~\cite{barron2021mip,DBLP:journals/corr/abs-2107-02791,roessle2022dense}, accelerate training and inference~\cite{mueller2022instant,liu2020neural,DBLP:journals/corr/abs-2103-14645,chen2022tensorf,Chen2023FactorFA}, improve geometry \cite{Wang2021NeuSLN,yariv2021volume,Yu2022SDFStudio,Oechsle2021ICCV}, and learn with fewer viewpoints \cite{Niemeyer2021RegNeRFRN,Roessle2021DenseDP,uy-scade-cvpr23}. In this work we are most interested in related research that integrates \textit{segmentation} into NeRF~\cite{kobayashi2022distilledfeaturefields,zhi2021inplace,kerr2023lerf,xu2023jacobinerf,liu2023instance,liu2022unsupervised,mirzaei2023spinnerf,yin2023ornerf,cen2023segment,KunduCVPR2022PNF,Siddiqui2022PanopticLF}.

One line of research aimed to learn a semantic or instance branch alongside the NeRF's color and density, thus yielding view-consistent segmentation \cite{zhi2021inplace,liu2023instance,vora2021nesf,KunduCVPR2022PNF,Siddiqui2022PanopticLF}. Semantic-NeRF~\cite{zhi2021inplace} was a seminal work that learned a semantic field to propagate 2D segmentations into new views. Instance-NeRF~\cite{liu2023instance} improved Semantic-NeRF by handling panoptic segmentation from Mask2Former~\cite{cheng2022masked}. Nevertheless, these methods are restricted to a closed vocabulary of instance categories, and additionally do not attempt a hierarchical understanding of the segmented instances.

Another recent direction has been to learn a generic volumetric feature field alongside NeRF. DFF~\cite{kobayashi2022distilledfeaturefields}, N3F~\cite{tschernezki2022neural}, and LeRF~\cite{kerr2023lerf} distilled 2D image features generated by off-the-shelf feature extractors such as CLIP~\cite{radford2021learning}, DINO~\cite{caron2021emerging}, and LSeg~\cite{li2022languagedriven} into feature fields that enable 3D segmentation and editing in NeRF. However, these methods often struggle to recover precise segmentation boundaries, and also do not establish a hierarchical structure on the features. In contrast, we learn a feature field that distills a hierarchy of segmentations from noisy multi-view SAM predictions.

Finally, other works~\cite{yin2023ornerf, cen2023segment, cen2024segment3dgaussians} used NeRF or 3D Gaussians~\cite{kerbl3Dgaussians} to propagate a single SAM query into novel views and establish view-consistency. However, these methods handle one segmentation at a time, while our approach trains a feature field to jointly aggregate and reconcile \textit{hundreds} of noisy 2D masks.

\subsubsection{Hierarchical Segmentation}
Hierarchical segmentation, a specialized domain within image segmentation, partitions an image into regions that exhibit a hiearchical tree structure, (i.e., each segmentation can be recursively divided into smaller segmentations). In the pre-deep learning era, researchers utilized non-parametric approaches to generate contours and enable hierarchical segmentation~\cite{1544874,beucher1994watershed,najman1996geodesic,1640630,arbelaez2010contour}.  Ultrametric distance and the watershed transform were employed to identify hierarchical clusters based on RGB values~\cite{najman1996geodesic,beucher1994watershed,1640630}, and subsequent advancements by Yarkony et al.~\cite{NIPS2015_3416a75f} and Xu et al.~\cite{xu2016hierarchical} improved the efficiency and flexibility of ultrametric hierarchical segmentation methods. More recently, Zhao et al.~\cite{zhao2017open} and Li et al.~\cite{li2022deep} shifted their focus from hierarchical segmentation \textit{within} an image toward estimating the hierarchy of segmentation 
\textit{classes}. This class hierarchy has also been of particular interest in the field of human parsing, where segmentation is performed according to the hierarchical structure of the human body~\cite{de2008hierarchical,wang2019learning,wang2020hierarchical,gong2019graphonomy}. 
In the 3D domain, Mo et al.~\cite{Mo_2019_CVPR} introduced PartNet, a large scale 3D mesh dataset annotated with hierarchical segmentation, and researchers have subsequently explored hierarchical structure in 3D shapes, including  multi-granularity segmentation on point clouds~\cite{sun2022semantic, sun2022semisupervised, zhou2023partslipenhancinglowshot3d}.

Most relevant to us, the recent Segment Anything Model~\cite{kirillov2023segment} emphasized \textit{multi-granular} segmentation in its open-vocabulary and zero-shot segmentation setting. Given its impressive performance, we use off-the-shelf SAM segmentations as supervision for our method. Furthermore, while concurrent works \cite{garfield2024, ying2023omniseg3d} also attempt to distill SAM masks into NeRF, our ultrametric feature field constitutes a fundamentally different approach to the problem.

\section{Preliminaries and Notation}
\label{sec:background}

In this section we describe the core  building blocks of our approach: (1) implicit feature fields, (2) hierarchical segmentation via watershed transform and ultrametrics, and (3) an off-the-shelf image segmentation model.  

\subsubsection{NeRF and Feature Fields}

A Neural Radiance Field (NeRF)~\cite{mildenhall2020nerf} is a volumetric representation that outputs a density $\sigma$ and color $\textbf{c}$ given a 3D coordinate $\textbf{x} = (x,y,z)$ and 2D viewing direction $\textbf{d}$. Given a pixel's camera ray $\textbf{r}$, the NeRF samples $N$ points along the ray $\textbf{x}_1, ..., \textbf{x}_N$ at corresponding intervals $\delta_1, ..., \delta_N$ and performs approximate volume rendering to estimate the pixel's color:
$    \textbf{C}(\textbf{r}) = \sum_{k=1}^{N} \rm T_k (1-e^{(-\sigma_k\delta_k)})\textbf{c}_k\,,$
%\end{equation}
with $\rm T_k = e^{-\sum_{i=1}^{k}\sigma_i\delta_i}$.

Recently, Distilled Feature Fields (DFF)~\cite{kobayashi2022distilledfeaturefields} proposed to learn a volumetric feature fields within a NeRF. Given 3D coordinate $\textbf{x}$, DFF outputs a feature $\textbf{f}$ in addition to the original density $\sigma$ and color $\textbf{c}$. Using the same volume rendering equation, DFF renders a features in addition to colors. 
We follow DFF and learn a NeRF with an accompanying feature field.

\subsubsection{Watershed Transform and Ultrametrics}

The Watershed transform~\cite{beucher1994watershed,najman1996geodesic} is a traditional hierarchical segmentation method. The method interprets edge energies as a heightmap, and % provided with an image, the algorithm 
initiates a flooding process, 
wherein energy basins beneath a given threshold are merged into regions, and higher thresholds cause regions to merge. Since the ``water level'' is uniform across the whole space, this yields a hierarchical segmentation. 

\begin{figure}[t]
    \begin{subfigure}[b]{0.47\textwidth}
    \centering
        \includegraphics[height=0.75in]{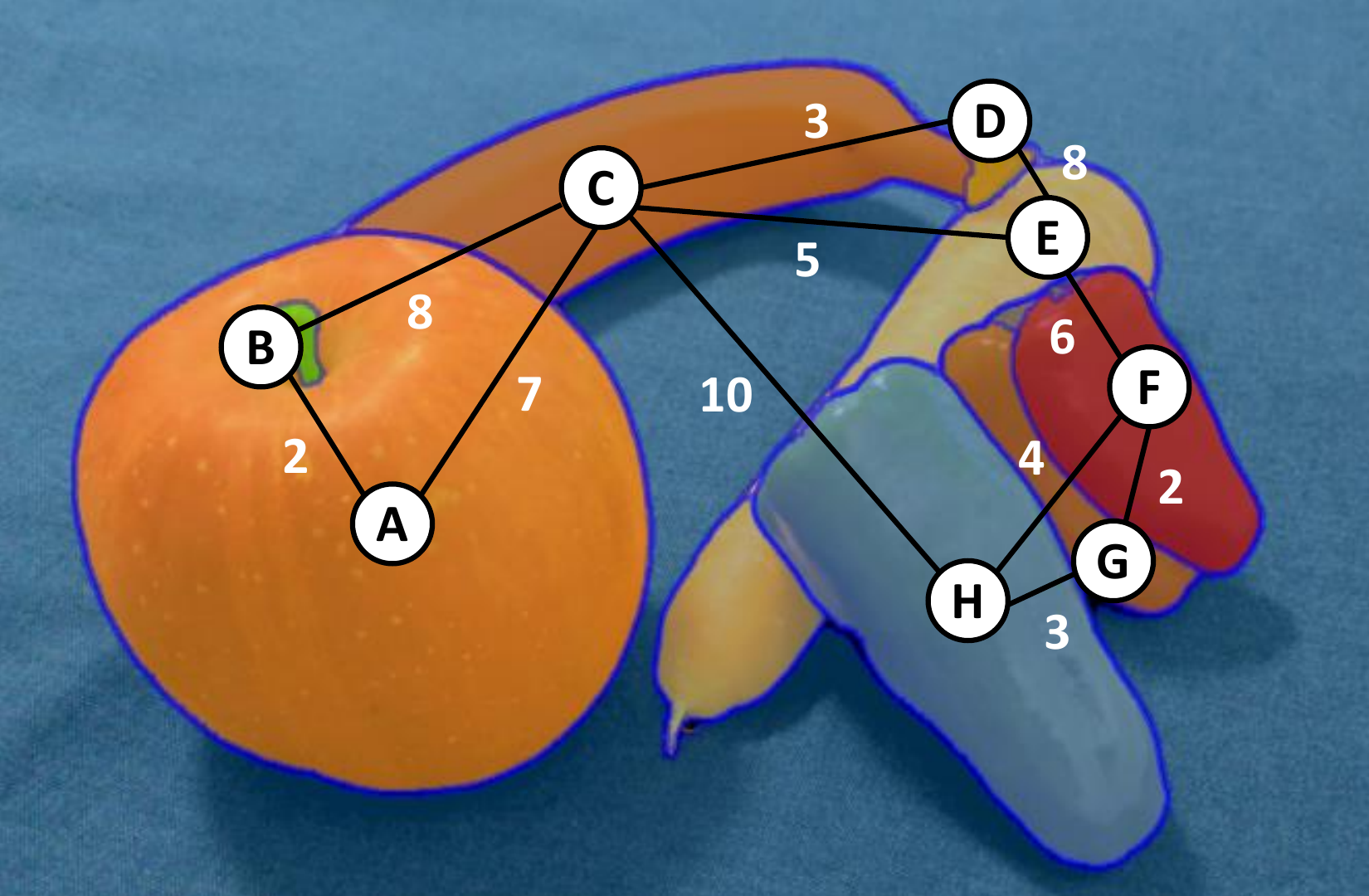}
    \end{subfigure}%
    ~
    \begin{subfigure}[b]{0.47\textwidth}
    \centering
        \includegraphics[height=0.75in]{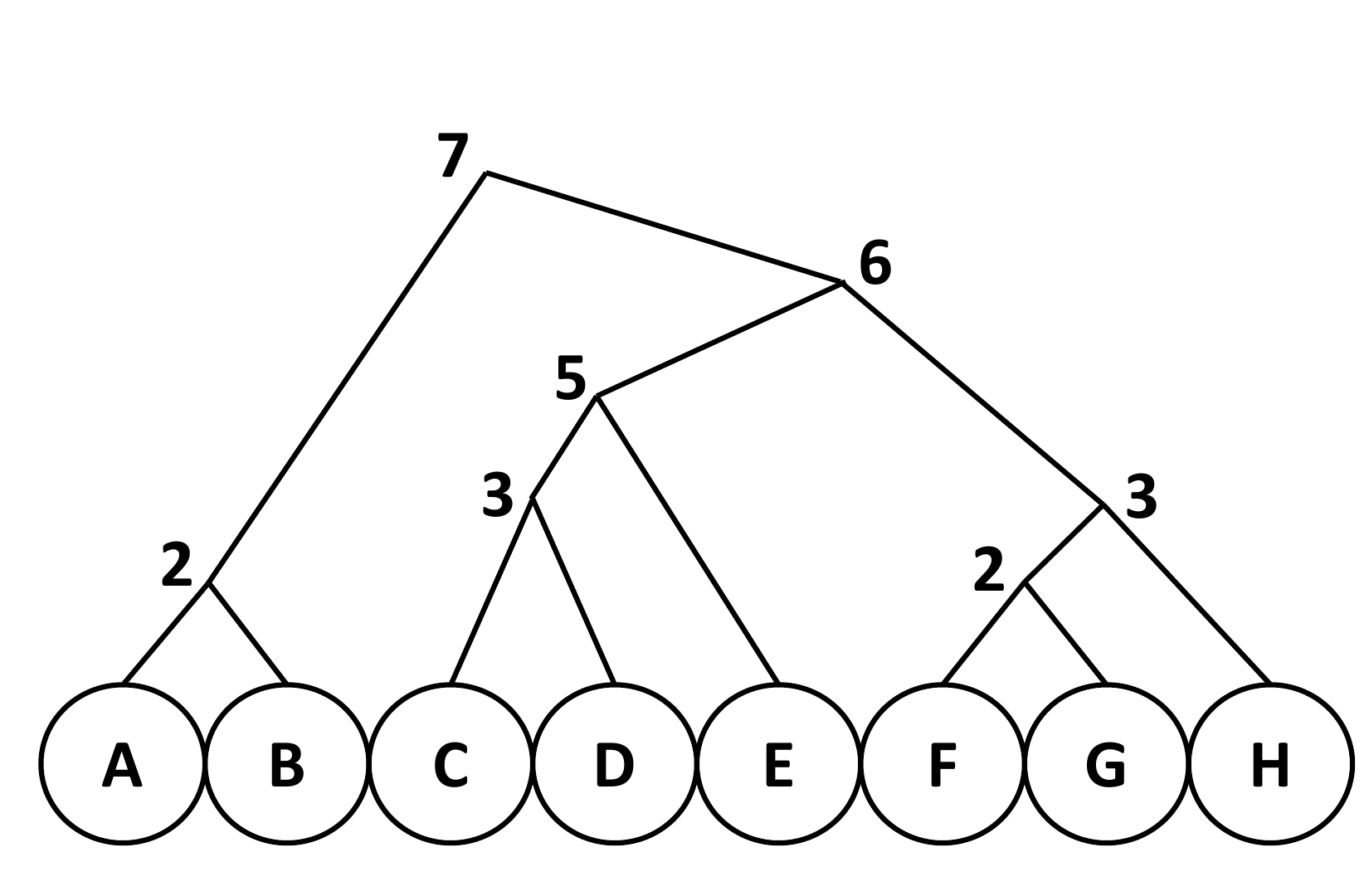}
    \end{subfigure}
    \caption{\textbf{Ultrametric Segmentation}: \textit{Left}: We overlay a simple graph on the image, showing edge lengths corresponding to feature distances between points. \textit{Right}: The hierarchical segmentation derived from the graph on the left. The numbers on the tree indicate the ultrametric distance between nodes on the two branches.
    }
    \label{fig:ultrametric}
    % \vspace{-2mm}
\end{figure}

Representing an image or scene as a graph, denoted as $\mathcal{G} = (V, E)$, where $V$ includes all of the pixels/points and $E$ connects points which are adjacent, the minimum water level that merges two points can be expressed as: 
\begin{equation} \label{eq:ultrametric}
    d(v_i, v_j) = \min_{p\in P} \max_{e \in p} |e|\,,
\end{equation}
where $P$ denotes all paths that connect $v_i$ and $v_j$ in the graph, and $e$ is an edge on the path $p$. 
Computing the distance between $v_i$ and $v_j$ means first finding the shortest path between the nodes, where path length is determined by the maximum edge along the path, and then reporting that crucial edge length. This is sometimes  called the minimax path problem. 
This distance is an \textit{ultrametric} distance~\cite{milligan1979ultrametric,johnson1967hierarchical}, which satisfies a triangle inequality of the form 
\begin{equation}
    d(x,y) \leq \max \{d(x,z), d(y,z) \}\,.
    \label{eq:triangle}
\end{equation}
\cref{fig:ultrametric} provides an illustration of paths and distances in a simple scene graph.

In our setup, we use \cref{eq:triangle} to define an ultrametric contrastive loss for our scene features. We use feature distances as edge lengths in \cref{eq:ultrametric}, enabling us to obtain hierarchical segmentations at test time via the Watershed transform.

\subsubsection{Segment Anything Model (SAM)}
SAM~\cite{kirillov2023segment} is a state-of-the-art vision foundation model for image segmentation.
Given a query in the form of points, a mask, a box, or a language prompt, SAM predicts a segmentation that best reflects the prompt. SAM generates segmentations at three levels of granularity: instance, part, and subpart. 

While SAM produces excellent results on \textit{single images}, it is nontrivial to lift its predictions into 3D. This is because SAM's predictions 
are not consistent across viewpoints. 
% View-dependent effects such as occlusion, lighting, camera zoom, and camera position can influence the regions that SAM segments, and also influence which ``granularity" SAM assigns.
For example, a pen may be segmented into multiple parts in a close-up view, then segmented as a single instance in a wider view, and then be completely missed when viewed from farther away. 
Furthermore, 
it is unclear how to best lift outputs from a ``queryable" design into a well-organized 3D representation: given an abundance of queries, SAM will generate an abundance of masks, which overlap with one another unpredictably, and have no straightforward unification. 

In our setup, we use SAM to provide a noisy signal of viable segmentations within each viewpoint, and rely on multi-view feature field optimization, using ultrametrics, to distill this knowledge into a 3D scene segmentation which is view-consistent and hierarchical.

\begin{figure*}[t!]
    % \vspace{-2mm}
    \centering
    \begin{subfigure}[t]{0.23\textwidth}
        \centering
        \includegraphics[height=0.65in]{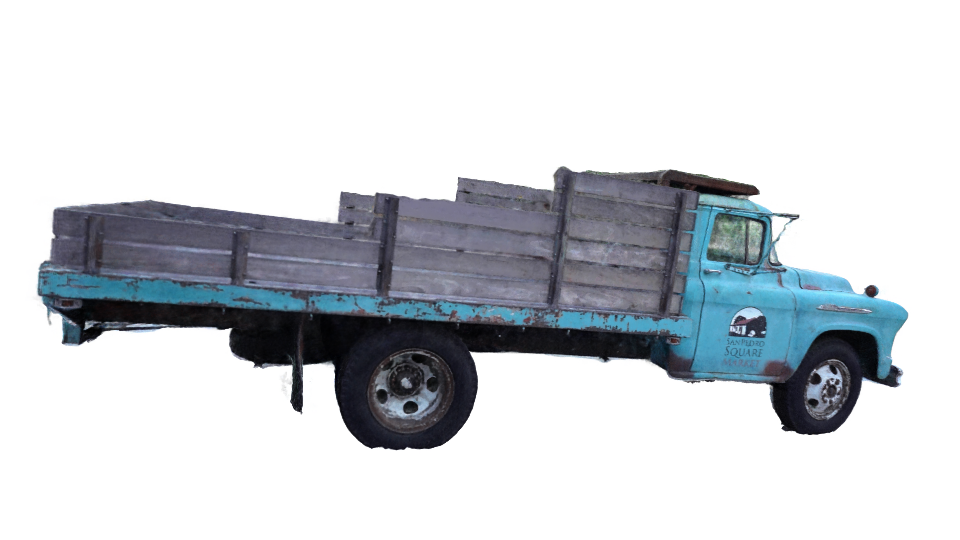}
        %\caption{Image}
    \end{subfigure}%
    \begin{subfigure}[t]{0.23\textwidth}
        \centering
        \includegraphics[height=0.65in]{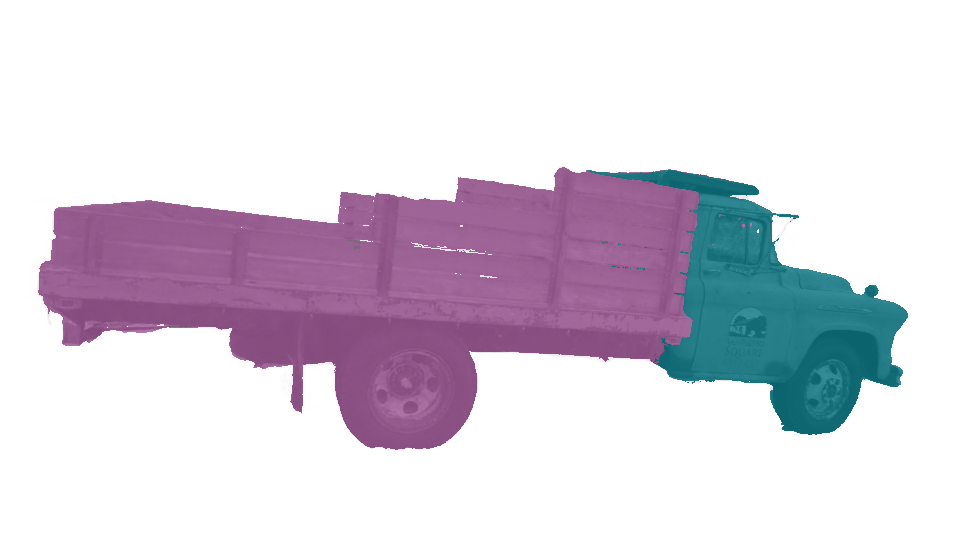}
    \end{subfigure}%
    \begin{subfigure}[t]{0.23\textwidth}
        \centering
        \includegraphics[height=0.65in]{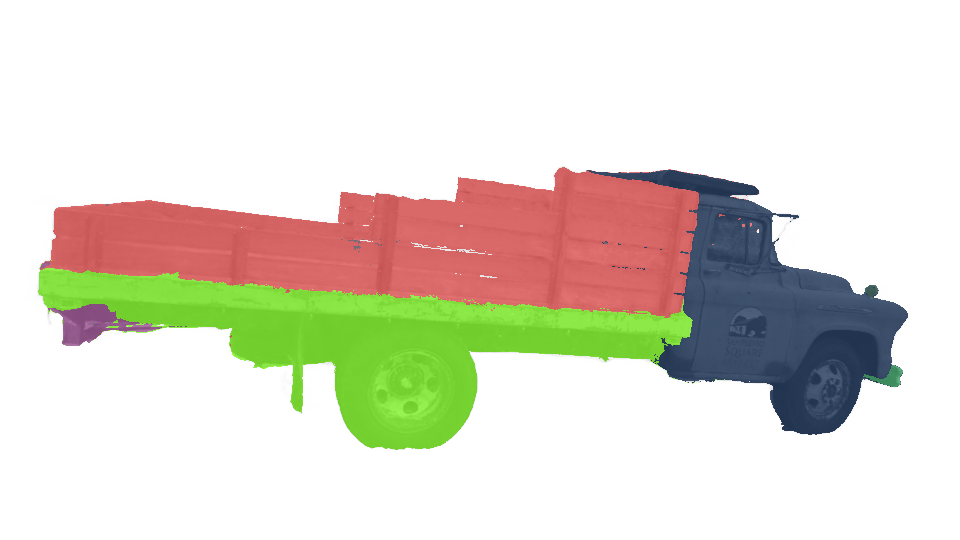}
        %\caption{Hierarchical Segmentation}
    \end{subfigure}%
    \begin{subfigure}[t]{0.23\textwidth}
        \centering
        \includegraphics[height=0.65in]{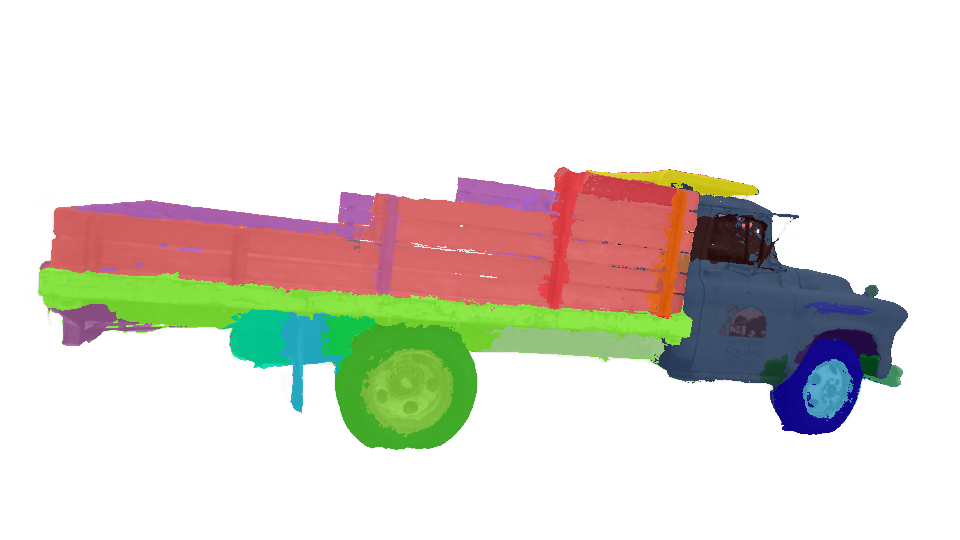}
    \end{subfigure}
    % \vspace{-2mm}
    \caption{\textbf{Hierarchical Segmentation}: Our method can hierarchically segment real world scenes at various levels of granularity.}
    \label{qual-seg-truck}
\end{figure*}

\section{Method}
\label{sec:method}

The previous section described the core components of our approach. In this section we describe how these pieces fit together, and explain the training and inference pipelines. 

\subsection{Learning an Ultrametric Feature Field} \label{sec:image-segmentations}

%\subsubsection{Problem Formulation}

\subsubsection{Problem Formulation} 
We take as input a set of multi-view images, along with their camera parameters. We run SAM on every image, typically yielding 50-150 masks per image. 
Our goal is to learn an implicit 
feature field that encodes a hiearchical understanding of these masks. 

\subsubsection{Contrastive Learning}
The SAM masks are inconsistent across views, but carry a great deal of information within each view. We use contrastive learning ~\cite{khosla2020supervised,chen2020simple,radford2021learning} to distill this information into a 3D feature field. 

We sample a pair of pixels in an image, and define it as a positive pair if both lie within a same mask, and a negative pair otherwise. We supervise the features of positive pairs to be more similar than those of negative pairs. Concretely, we follow Chen et al.~\cite{chen2020simple} and minimize a binary cross entropy loss on distances between positive pairs and negative pairs. Given a positive pair $s_p = \{ v_{p1}, v_{p2}\}$ and a negative pair $s_n = \{ v_{n1}, v_{n2}\}$, our contrastive loss is 
\begin{equation}
\label{eq:contrastive}
\begin{split}
    \ell(s_p, s_n) & = -\log(\frac{e^{d(v_{p1}, v_{p2})/\tau}}{e^{d(v_{p1}, v_{p2})/\tau} + e^{d(v_{n1}, v_{n2})/\tau}})\\
    &+ \log(\frac{e^{d(v_{n1}, v_{n2})/\tau}}{e^{d(v_{p1}, v_{p2})/\tau} + e^{d(v_{n1}, v_{n2})/\tau}})\,,\\
\end{split}
\end{equation}
where $d$ is a distance metric, and $\tau$ is the temperature.

Minimizing the loss in \cref{eq:contrastive} across many pairs $s_p$ and $s_n$ yields a feature space where segmentations can be recovered by querying a point and finding the neighborhood where pairwise feature distances are all below some threshold. As discussed in earlier sections, whether or not the resulting segmentation will be hierarchical depends on whether the distance metric $d$ is an ultrametric or a standard Euclidean metric.
Our main goal is to reduce ultrametric distances, but to improve optimization, we apply loss in both ultrametric space and Euclidean space:
\begin{equation}
\label{eq:loss}
    L_{feat} = \sum_{(s_p, s_n) \in \mathcal{S}} \ell_{\textrm{ultra}}(s_p, s_n) + \alpha \ell_{\textrm{Euclid}}(s_p, s_n)\, .
\end{equation}
The ultrametric term forces the optimization to find a hierarchical decomposition of the scene, while the Euclidean term serves as regularization. Note that the Euclidean term has the advantage of directly providing a gradient to all considered feature pairs, whereas the ultrametric loss only provides a loss to the active (maximal) edge along the path between two segments. 

As described in \cref{eq:ultrametric,eq:triangle}, ultrametric distances are defined on a graph. At each training step, we form an approximate graph by sampling 4096 pixels within the image, connect each pixel to its 10 nearest neighbors, and use feature distances as edge weights. 
% Minimizing the ultrametric distance of a positive pair $s_p$ means finding the shortest path between the vertices, and penalizing the longest edge on that path. 
We use the binary partition tree algorithm~\cite{cousty2018hierarchical} to efficiently compute ultrametric distances during training. 

\subsubsection{Hierarchical Sampling}
\label{sec:hier-sampling}
Because our input segmentations overlap within and across views, it is ambiguous whether a pair of pixels ``lie within the same mask'' and therefore represent a positive pair in our contrastive formulation. For example, two pixels may be within the same mask at a coarse granularity but not for a finer granularity. We address this ambiguity with a simple tree-based strategy for sampling positive and negative pairs of pixels.  % , which is presented in the supplementary. % \cref{alg:data_sampling}.

For each image, we organize the segmentation masks into a hierarchical structure determined by the inclusion ratio between them. We additionally include an all-positive mask as the root of the tree. 
Then, for each view, we sample positive and negative pairs starting from the leaf nodes of the hierarchical tree, \ie masks at the finest granularity. For a leaf mask $A$ with parent mask $B$, we randomly select two pixels in $A$ and designate these as a positive pair. We then randomly select one pixel in $A$ and one pixel in $\bar{A} \cap B$ and designate these as a negative pair. We then move to the parent mask and repeat.
% We sample an equal number of positive and negative pairs. 
Additional details are provided in the supplementary.

We note that masks grow in size as we rise up the hierarchy, and pairs which were declared negative at finer granularity will be declared positive in courser segmentations. We mitigate this conflict by computing the contrastive loss \textit{within each level}.
% and summing the result across levels
 This asks that the positive distances be smaller than the negative distances within a single level, and avoids the possibility of a pair being simultaneously positive and negative in \cref{eq:contrastive}.
Taken together, our sampling strategy reflects an ultrametric structure and helps the segmentation hierarchy propagate across the scene.

\begin{comment}
\begin{algorithm}[t]
\caption{Data Sampling}
\label{alg:data_sampling}
\begin{algorithmic}
    \STATE pos\_samples $\gets$ []
    \STATE neg\_samples $\gets$ []
    \FORALL{$A$ $\in$ leaf\_masks}
        \STATE  \# sample positive pairs from the leaf node
        \STATE sample $\gets$ (Random($A$), Random($A$)) 
        \WHILE{$A$ has Parent}
            \STATE $B$ $\gets$ $A$.Parent
            \STATE pos\_samples += sample
            \STATE \# sample negative pairs for the current level
            \STATE sample $\gets$ (Random($A$), Random($\bar{A} \cap B$))
            \STATE neg\_samples += sample
            \STATE $A$ $\gets$ $B$            
        \ENDWHILE 
    \ENDFOR
    \RETURN pos\_samples, neg\_samples
\end{algorithmic}
\end{algorithm}
\end{comment}

\subsubsection{Improving Depth with Segmentation}
Our approach shows that the 3D structure of a scene can resolve conflicts in segmentation cues. Conversely, can segmentation cues help resolve ambiguity in 3D structure? 
To explore this possibility, we add an assumption that regions belonging to the same semantic mask have smoothly changing depth. We propose a regularization that penalizes changes in curvature (\ie the third derivative of depth) within a segment. Concretely, at each training iteration, we sample $k_{depth}$ local patches of $4 \times 4$ pixels, ensuring that each sampled patch lies within one of the SAM's finest-grained mask predictions. The depth continuity loss is defined as
\begin{equation}
    L_{dc} =\sum_{p_0\in P_D}{\max(\frac{(d_{p_0}-3 d_{p_1}+3d_{p_2} - d_{p_3})}{( \max(d_{p_0}, d_{p_1}, d_{p_2}, d_{p_3})\Delta \theta)^ 3} - t, 0)}
\end{equation}
where $\Delta \theta$ represents the ray angle difference between adjacent pixels, $t$ denotes a threshold, and $P_D$ contains the sampled pixels, with $p_0, \ldots, p_3$ denoting adjacent pixels in a row or column. For training stability, we start using this loss halfway through training. \cref{fig:depth continuity} presents the benefits of the depth continuity loss.
% our depth contia visual comparison of depth estimation with and without the depth continuity loss.

\subsection{Segmentation from Ultrametric Features}
\label{sec:ultrametric-eval}

After training our ultrametric feature field, we can perform 2D or 3D hierarchical segmentation by applying the Watershed transform on either rendered feature maps or the 3D feature field itself.

\subsubsection{Segmenting in 2D}

To segment a 2D image from our feature field, we begin by rendering our feature field to the viewpoint of the image. We then construct a graph from the feature map, denoted as $\mathcal{G} = (V, E)$, where $V$ contains the pixel features, and $E$ contains edges connecting each pixel to its 4 spatial neighbors. Edge lengths are defined as feature distances. With a given threshold $t$ as the indicator of granularity, we remove all edges longer than this threshold, resulting in a new graph denoted as $\mathcal{G}_t$. We then identify all connected components within this graph, and return these components as our segmentation. 
We use this approach to compare against 2D segmentation methods, but we note that 2D-based segmentation is not view-consistent. 

\subsubsection{Segmenting in 3D}
To achieve 3D-consistent segmentation, we create a featurized 3D point cloud by unprojecting feature maps and depth maps rendered from the optimized implicit field. Following related NeRF works, we remove outliers and downsample the pointcloud using Open3D~\cite{Zhou2018}.
Then, we construct a k-nearest-neighbor (KNN) graph on the 3D point cloud, and we set each edge weight in the KNN graph to the feature distance between the connected vertices. 
For a threshold $t$ indicating the level of granularity, we remove all edges longer than that threshold, and the remaining connected components represent segments at the specified level of granularity. 
% Since this process occasionally yields tiny unusable segments, 
In practice, we keep the $N$ largest components as our final segmentation. 
Using different values of $t$ allows for segmentation at varying levels of granularity.

To propagate the 3D segmentation into a novel view, we first render a depth map of the novel view and unproject the render into 3D points. Then, for each unprojected pixel, we find its $k$ nearest neighbors to the previously estimated KNN graph and assign a segmentation label using the mode of the neighborhood. 

\begin{figure}[t!]
    \centering
    \begin{subfigure}[t]{0.23\linewidth}
        \centering
        \includegraphics[height=0.76in]{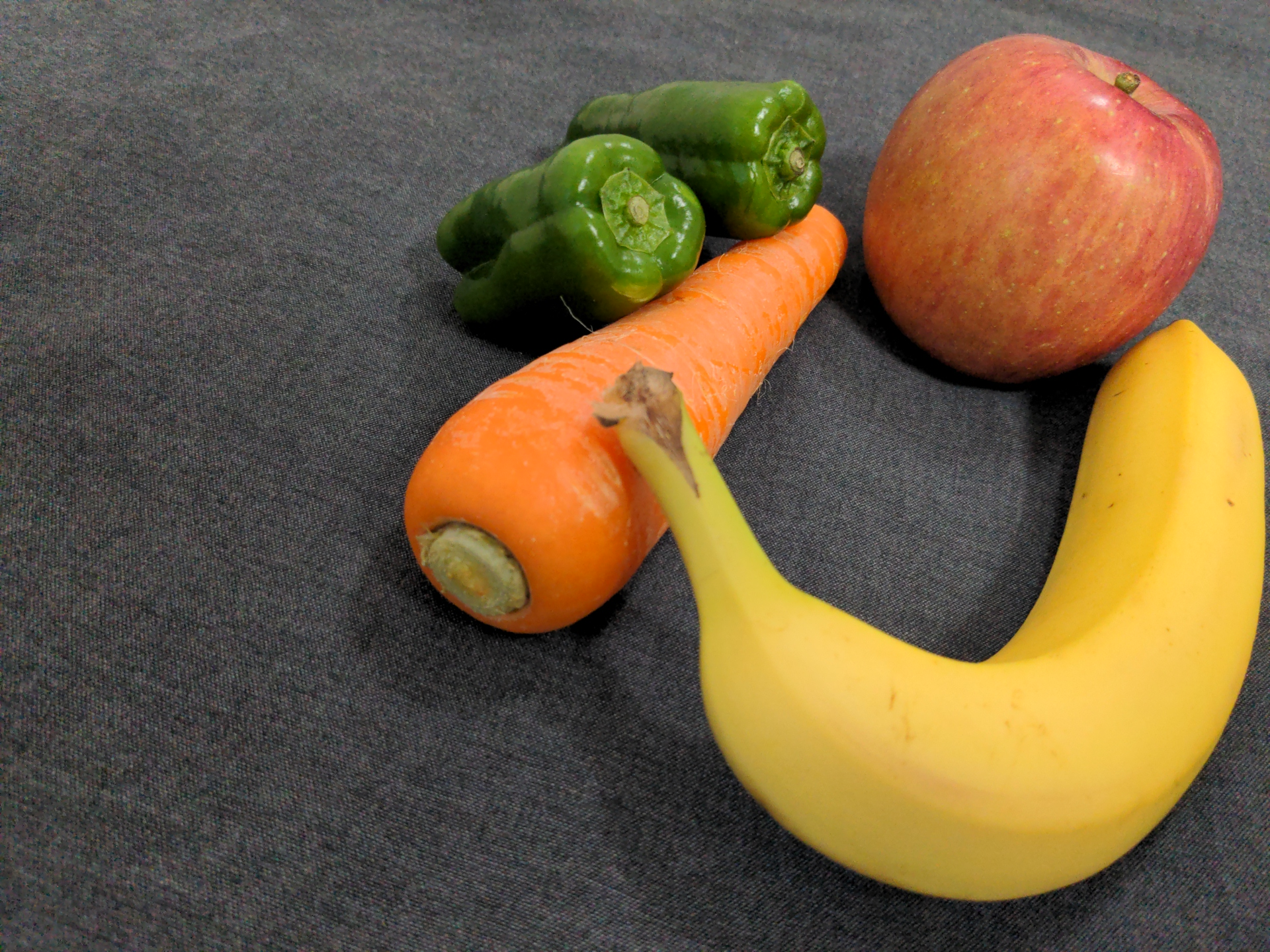}
        \caption{Image}
    \end{subfigure}
    \hfill
    \begin{subfigure}[t]{0.23\linewidth}
        \centering
        \includegraphics[height=0.76in]{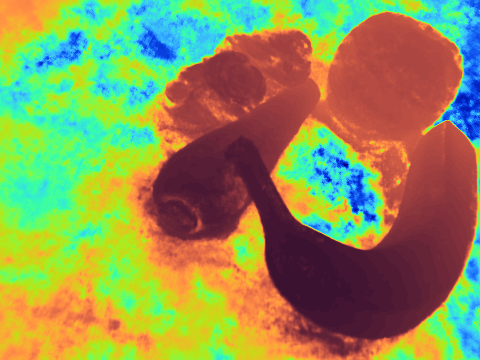}
        \caption{Original}
    \end{subfigure}
    \hfill
    \begin{subfigure}[t]{0.23\linewidth}
        \centering
        \includegraphics[height=0.76in]{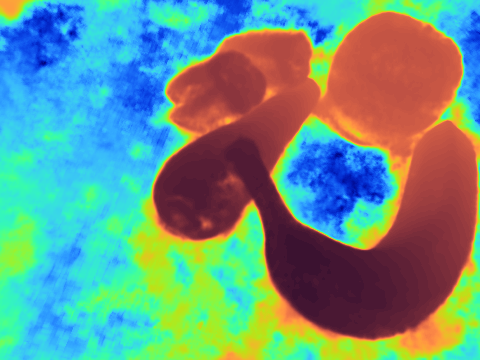}
        \caption{DC}
    \end{subfigure}
    \hfill
    \begin{subfigure}[t]{0.23\linewidth}
        \centering
        \includegraphics[height=0.76in]{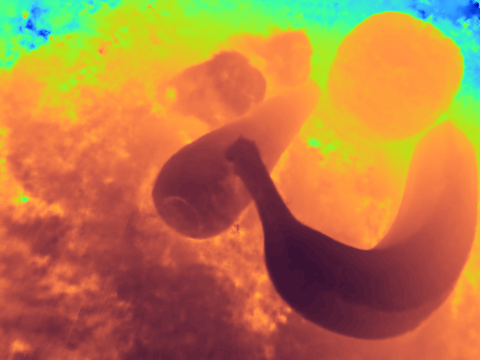}
        \caption{DC \& COLMAP}
    \end{subfigure}
    % \vspace{-2mm}
    \caption{\textbf{Depth Continuity}: Our depth continuity loss (labeled as DC above) leads to smoother and more plausible depth estimation, and can be seamlessly combined with additional depth cues such as COLMAP~\cite{schoenberger2016mvs}, resulting in even greater accuracy.
    }
    % \vspace{-3mm}
    \label{fig:depth continuity}
\end{figure}

\section{Experiments}
\label{sec:exp}
We report qualitative and quantitative evaluations for our method and a set of relevant baselines. 
In \cref{sec:dataset} we give an overview of the datasets we evaluate on, and we define our evaluation metrics. 
In \cref{sec:comparison} and \cref{sec:ablation}, we present quantitative comparisons and ablation studies.
% We additionally visualize our method's hierarchical and view-consistent segmentation on real-world scenes from the Tanks and Temples Dataset~\cite{Knapitsch2017}.
We present implementation details in \cref{sec:implmentation}.

\subsection{Datasets and Metrics}
\label{sec:dataset}

\paragraph{PartNet Dataset} The official PartNet dataset ~\cite{Mo_2019_CVPR, chang2015shapenet} contains $26,671$ 3D models with professionally verified hierarchical part decomposition. Furthermore, each object model is rendered into 24 viewpoints that uniformly cover the viewing sphere. PartNet offers us the unique ability to evaluate on hierarchical segmentation that verifiably aligns with human perception of structural hierarchy. We evaluate on five objects from each of the \textit{chair}, \textit{table}, and \textit{storage furniture} categories, totalling to 15 objects models. 

\paragraph{Blender with Hierarchical Segmentation}
While PartNet contains 3D objects with hierarchical part annotations, each model is visually simplistic and lacks the photometric complexity often associated with NeRFs. 
To the best of our knowledge, no publicly available dataset exists that displays both complex photometric structure as well as 3D-consistent hierarchical segmentation labels. Thus, we create a new synthetic dataset based on the Blender Dataset~\cite{mildenhall2020nerf} which we call Blender with Hierarchical Segmentation (Blender-HS). 
We choose three scenes from the Blender Dataset that exhibit clear hierarchical structure: Lego, Hotdog, and Drums, and for each we define three granularity levels, which we denote as ``scene'', ``collection'', and ``object''. The object level consists of each distinct 3D asset; the ``collection'' level contains 3D asset groupings that were decided by the original artist; the ``scene'' level consists of the entire scene. 
While many potential hierarchies exist in general, the artist-defined groupings naturally represent a hierarchy that makes sense for humans.  
We present visualizations of the hierarchical ground truth in the supplementary. We use 100 views in the training set for training and 10 views from the validation set for evaluation.

\noindent\paragraph{Normalized Covering Score}

To assess the quality of hierarchical segmentation, we use the Normalized Covering (NC) score~\cite{ke2022unsupervised}. This metric averages the Intersection over Union (IoU) between each ground truth mask and the \textit{best-matching} (\ie most-overlapping) predicted mask. 
We calculate a separate NC score for each granularity and report the mean of these scores. We also adapt the metric to evaluate the accuracy of point clouds segmentation on PartNet~\cite{Mo_2019_CVPR}.

\noindent\paragraph{Segmentation Injectivity Score}
In hierarchical segmentation, it is important that each pixel belongs to only one mask for each level of granularity. To measure this, we propose a metric which we call Segmentation Injectivity (SI). 
For each ground truth segmentation, we retrieve its {best-matching} predicted mask and sample two random points $p_1$ and $p_2$ from that mask. We then query the model for a new mask at each of these points, at the same granularity. We compute the Intersection over Union (IoU) between the two resulting masks.  %to predict new masks centered about $p_1$ and $p_2$. 
Note that a perfect model will return the same mask at both locations, since these points belong to the same ground-truth segment, whereas a worse model will return masks which may not even overlap. Since the random point sampling introduces randomness, we run this 100 times per ground truth mask and average the scores across all ground truth masks and all viewpoints. We find this is more tractable than estimating masks densely, and in practice exhibits low variance. 
Note there is a trade-off between the SI score and the NC score: predicting an excessive number of overlapping masks could inflate the NC score but would significantly lower the SI score. 

\noindent\paragraph{View Consistency Score}

We use a View Consistency (VC) score to measure the 3D consistency of image segmentations. The key idea of this score is to estimate segmentations in two nearby viewpoints independently, and then warp these estimates onto one another, and measure their agreement. 
To compute this score, we begin by defining a pixel transformation $T$ from a source viewpoint to another one shifted 10 degrees, using ground truth depth and camera pose.
Next, for each ground truth segmentation, we retrieve its best-matching predicted mask, sample a random point $p_1$ from that mask, and compute the point's location in the shifted viewpoint $p_2=T(p_1)$. We then obtain segmentation estimates to compare, centered on $p_1$ in the first view and centered on $p_2$ in the second. 
Finally we warp $p_2$'s mask into the $p_1$ viewpoint, and measure IoU with the $p_1$ mask. To ensure that the IoU does not merely reveal occlusion/disocclusion differences, we use ground truth visibility masks to remove pixels which are invisible in either view, before computing the IoU. 
Similar to the other scores, we 
% mitigate noise induced by the random sampling by computing
compute this score 100 times per ground truth mask and average the scores across all ground truth masks and all viewpoints.
\subsection{View-Consistent Hierarchical Segmentation}
\label{sec:comparison}

\begin{table}[tb]
  \caption{\textbf{Results on Blender-HS.} We report the Normalized Covering (NC), Segmentation Injectivity (SI), and View Consistency (VC) as a percentage. NC$_\text{obj}$, NC$_\text{coll}$, and NC$_\text{scene}$ refer to the NC score on objects, collections, and scenes, respectively.
  }
  % \vspace{-2mm}
  \label{tab:comparison}
  \centering
  \resizebox{0.75\linewidth}{!}{
  \begin{tabular}{@{}lcccccc@{}}
    \toprule
    %Heading level & Example & Font size and style\\
    Method & NC$_{obj}$ $\uparrow$ & NC$_{coll}$ $\uparrow$ & NC$_{scene}$ $\uparrow$ & NC$_{mean}$ $\uparrow$ & SI $\uparrow$ & VC $\uparrow$ \\ %\cline{2-5} 
    \midrule
   LSeg~\cite{li2022languagedriven} &21.8&	40.6&	67.0&	43.2 & 69.8 & 53.6\\
   LSeg + DFF~\cite{kobayashi2022distilledfeaturefields} &16.4&	42.7&	83.9	&47.7 & 84.8 & \textbf{82.7}\\
   LeRF~\cite{kerr2023lerf} &26.6 & 41.3	&85.7&	51.2 & 59.8 &64.1  \\
   SAM~\cite{kirillov2023segment} &44.3 &	65.2 &78.7  & 62.7 & 80.8 & 46.2 \\
   SAM3D~\cite{cen2023segment} & 46.0	&59.3	&53.2	&52.8 & - & 72.4 \\
   \hline
   Ours, 2D &\textbf{50.0}	&63.0&	94.0	&69.0 & \textbf{100.0}  & 67.8\\
   Ours &48.0	&\textbf{67.7}&	\textbf{97.1}&	\textbf{70.9} & \textbf{100.0} & 78.9 \\
  \bottomrule
  \end{tabular}
  }
  % \vspace{-3mm}
\end{table}

\paragraph{Baselines}
Because the exact task of 3D-consistent hierarchical segmentation from images collections is less explored, we adapt state-of-the-art NeRF baselines to this setting. We adapt language-based methods LSeg~\cite{li2022languagedriven}, DFF~\cite{kobayashi2022distilledfeaturefields}, and LeRF~\cite{kerr2023lerf} to predict a segmentation mask based on a 2D pixel query, by querying the rendered feature map and estimating a segmentation mask via feature similarity. We generate multi-granular predictions by using different thresholds on the feature similarity. 
We also adapt SAM3D~\cite{cen2023segment} as a baseline, which we modify to produce multiple segmentations instead of only one. We propagate SAM outputs from 20 training views into the novel view to generate a variety of masks.

\paragraph{Quantitative Evaluation}
In Table \ref{tab:comparison}, we report segmentation metrics for our 2D inference mode, our 3D inference mode, and our baselines.  
Our method significantly outperforms all other methods in Normalized Covering score (\ie segmentation accuracy). We even outperform SAM, suggesting that our ultrametric feature field not only distills thousands of SAM predictions into a compact set of hierarchical and view-consistent masks, but that our masks are more accurate than \textit{any} of the original SAM predictions.  
Due to the Watershed transform, we also achieve a perfect score on segmentation injectivity (SI), while the other methods (which are non-hierarchical) 
tend to predict overlapping masks. Finally, our 3D inference shows significant improvements in view consistency over all methods except DFF. We also point out that 
DFF's object-level NC score suggests that it does not segment small objects, which are significantly more challenging to segment in a view-consistent manner, and this fact may inflate the VC score for the method. 

We additionally visualize our ultrametric feature field on the Lego scene in the supplementary, 
showing sharper features than DFF.

\paragraph{Qualitative Analysis}
We provide qualitative validation of our method on two real-world scenes from the Tanks and Temples Dataset~\cite{Knapitsch2017}. In \cref{qual-seg-truck}, we visualize three segmentation granularities on the truck scene. In \cref{fig:teaser}, we show our segmentations on the Bobcat tractor scene, which resolves inconsistencies from the source SAM masks. Our method's segmentations tend to have sharp boundaries, and the hierarchical structure decomposes the truck and tractor into intuitive parts and subparts.

\paragraph{Partnet Experiments}

Each object in PartNet has a unique hierarchical part decomposition that may vary in both the number of parts and number of relationships between parts. To evaluate the part hierarchy using our NC metric, we categorize the part hierarchy of each object into three levels: finest, middle, and coarse. The finest level corresponds to the leaf level, while the coarse level represents the root level (i.e., full object), and the rest are the middle level.
In Table \ref{tab:partnet}, we compare our approach with LSeg+DFF \cite{kobayashi2022distilledfeaturefields} and SAM3D \cite{cen2023segment}, two other methods that produce 3D-consistent segmentation. Our method surpasses both in NC score, while also being the only method to produce hierarchical segmentation. We also report the NC Score based on point clouds and compare our method with 3D-PIS~\cite{sun2022semantic}, a supervised point clouds segmentation network. We can see that our method can achieve comparable NC Score on the finest level. Our method falls behind on the middle level, as the ground truth hierarchical tree is constructed semantically, 3D-PIS learns these semantics from supervision while ours does not.

\begin{table}[tb]
  \caption{\textbf{Evaluation in PartNet.} We report the Normalized Covering (NC) as a percentage on PartNet dataset. NC$_\text{fine}$, NC$_\text{mid}$, and NC$_\text{coar}$ refer to the NC score on three different hierarchy levels. $^\dag$ means the method is evaluated based on point clouds.
  }
  % \vspace{-2mm}
  \label{tab:partnet}
  \centering
  \resizebox{0.58\linewidth}{!}{
  \begin{tabular}{@{}lcccc@{}}
    \toprule
    Method & NC$_{fine}$ $\uparrow$ & NC$_{mid}$ $\uparrow$ & NC$_{coar}$ $\uparrow$ & NC$_{mean}$ $\uparrow$ \\ %\cline{2-5} 
    \midrule
   LSeg + DFF~\cite{kobayashi2022distilledfeaturefields} & 22.4 & 48.5 & 70.6& 47.2 \\
   SAM3D~\cite{cen2023segment} & \textbf{51.0}	&60.8	& 66.8 & 59.6 \\
   %\hline
   Ours &49.9&	\textbf{63.3}&	\textbf{82.8} & \textbf{65.3} \\
   \midrule
   Ours$^\dag$ & 48.9 &	60.4 & -	 & -\\
   3D-PIS$^\dag$~\cite{sun2022semantic} & \textbf{52.2} & \textbf{75.0}&	- & - \\
  \bottomrule
  \end{tabular}
  }
  % \vspace{-2mm}
\end{table}

\subsection{Ablation Experiments}
\label{sec:ablation}

\noindent\paragraph{Hierarchical Data Sampling}
In \cref{tab:ablation_1} we evaluate the influence of hierarchical data sampling for contrastive learning.
In the baseline approach, positive pairs are randomly sampled within each mask, and negative pairs are sampled with one point inside and one point outside. Results indicate that using the hierarchical sampling strategy improves the overall performance by $5.7$ points.

\noindent\paragraph{Ultrametric training}
In \cref{tab:ablation_1} we report the impact of the ultrametric loss, $\ell_{\textrm{ultra}}$. 
In the baseline method, we train the feature space using only the Euclidean loss, $\ell_{\textrm{Euclid}}$. 
Results show that the ultrametric training improves the segmentation performance by $2.1$ points.

\noindent\paragraph{Depth Continuity Loss}
\label{sec:depth_loss}
\cref{tab:ablation_2} shows the influence of using the depth continuity loss during training. In addition to evaluating the NC score, we evaluate depth estimation accuracy, using 
mean $\ell_2$ error. Incorporating the depth continuity loss improves both segmentation accuracy and depth accuracy. 
\cref{fig:depth continuity} qualitatively illustrates the benefit of the depth continuity loss in real data. 

\begingroup
\setlength{\tabcolsep}{3pt} % Default value: 6pt
\begin{table}[t]
\parbox{.55\linewidth}{
\centering
\caption{Ablation on Hierarchical Sampling (HS) and Ultrametric Training (UT).}
% \vspace{-2mm}
\resizebox{0.85\linewidth}{!}{
  \begin{tabular}{@{}lcccc@{}}
    \toprule
   {Method} & NC$_{\text{obj}}$  & NC$_{\text{coll}}$  & NC$_{\text{scene}}$ & NC$_{\text{mean}}$ \\
   \midrule
   Ours & \textbf{48.0}	& \textbf{67.7}	&\textbf{97.1}	&  \textbf{70.9}\\
   % \hline
   w/o UT &47.1	&62.8&	96.4	&68.8\\   
   w/o HS &42.6	&64.0	&89.0	&65.2\\        
  \bottomrule
  \end{tabular}
}
  \label{tab:ablation_1}
}
\hfill
\parbox{.4\linewidth}{
\centering
\caption{Ablation on Depth Continuity (DC) loss.}
% \vspace{-2mm}
\begin{tabular}{@{}lcc@{}}
\toprule
Method & NC$_{\text{mean}}$ & Depth Error  \\ 
\midrule
Ours &  \textbf{70.9} & \textbf{0.059}\\
% \hline
w/o DC &67.1 & 0.089\\
\bottomrule
\end{tabular}
\label{tab:ablation_2}
}
% \vspace{-3mm}
\end{table}

\endgroup

\subsection{Implementation Details}
\label{sec:implmentation}

We implement our method with help from the publicly available codebase from DFF~\cite{kobayashi2022distilledfeaturefields}. We model our feature branch after the RGB branch of Instant-NGP~\cite{mueller2022instant} -- we employ a multi-resolution grid hash encoder to transform 3D coordinates into features, followed by an MLP. For the grid hash encoder, we configure the number of levels to $17$, features per level to $4$, and the hash map size to $2^{20}$. The MLP comprises three hidden layers with $128$ dimensions each, producing a final output feature of $256$ dimensions. We provide additional implementation details in the supplementary.

For our quantitative evaluation on Blender-HS, we compute the NC score of LSeg, DFF, LeRF, and our method across 50 distance thresholds ranging from $0.01$ to $0.50$. We follow DFF~\cite{kobayashi2022distilledfeaturefields} and perform all evaluation on $4\times$ downsampled images. For all methods, we exclude masks containing fewer than $20$ pixels.

\section{Discussion and Limitations}

A key limitation of our approach is its dependence on high-quality point clouds produced by the NeRF% and the accuracy of 2D open-world segmentation
. While the depth-smoothing loss outlined in Section \ref{sec:depth_loss} improves point cloud quality, there is much room for further improvement. %We notice that there are some recent work that can improve the geometry estimation, and believe our method would benefit from these improvements.
% \adam{either cite it or nothing}
Regarding our evaluation, we find there is a scarcity of high-quality datasets for hierarchical 3D segmentation that exhibit complex scene structure. Additionally, it is ambiguous \textit{which} hierarchies are the most meaningful and appropriate in complex environments, without defining end-tasks that rely on these hierarchies. While our Blender-HS Dataset is a first step in providing ground truth on hierarchical 3D segmentation in complex scenes, it is limited to only three scenes. %For a more comprehensive evaluation, 
We hope that future efforts can develop better and larger hierarchical 3D datasets, to enable more comprehensive evaluations. 

\section{Conclusion}

Consistent hierarchical 3D segmentation is essential for many applications involving mobile agents interacting with the real world at scale, such as robotics and augmented reality. In this work we have demonstrated significant progress towards achieving consistent hierarchical segmentation, building on state-of-the-art systems that perform multi-granularity segmentation in images, whose output predictions are neither hierarchical nor consistent across different views. Our ultrametric feature field distills this inconsistent 2D information into a representation that can be queried at will for arbitrary-granularity segmentations that are consistent across views.

% \section{Conclusion}
% \label{sec:concl}
% For many applications that involve a mobile agent interacting with the real world at scale, such as robotics or augmented reality, a consistent hierarchical segmentation is essential. In this work we have demonstrated significant progress towards that goal, building on state-of-the-art systems that perform multi-granularity object and part detection in images, 
% whose output predictions are neither hierarchical nor consistent across different views. Our ultrametric feature field distills this inconsistent 2D information into a representation that can be queried at will for arbitrary-granularity segmentations that are consistent across views.
\paragraph{Acknowledgements}
This work was supported by a Vannevar Bush Faculty Fellowship and ARL grant W911NF-21-2-0104.

\bibliographystyle{splncs04}
\bibliography{refs}

\clearpage
\setcounter{page}{1}

\section{Supplementary Material}

We provide additional implementation details in Sec. A. 
We present additional qualitative results on Blender-HS dataset, PartNet dataset, and LLFF dataset in Sec. B. Additionally, we include a video attachment with visualizations of the view-consistent hierarchical segmentation results.

\subsection*{A. Additional Implementation Details}
\subsubsection{A.1. Hyperparameters}

On Blender-HS and PartNet, we train our model for $20,000$ iterations with a batch size of $4096$ and use the same optimization parameters as DFF. In contrastive learning (see \cref{eq:contrastive}), we set the temperature $\tau$ to $0.1$ and sample $64$ positive and negative pairs from each mask. We set the loss weight $\alpha$ of the Euclidean loss in \cref{eq:loss} to 1. For depth continuity loss, we sample $16$ patches per mask, and begin using the depth continuity loss after $5000$ iterations. 
During 3D inference, we extract a point cloud from training-view depth maps, apply voxel downsampling with voxel size $2\times 10^{-3}$, and run outlier removal with distance threshold $4\times 10^{-3}$ and number threshold of $1$. We build the graph of points using $k_{graph}=16$ nearest neighbors, and we transfer point segmentation labels into a novel view using the mode of $k_{query}=5$ nearest neighbors. We retain $N=200$ graph components and set the distance threshold $d$ to be $5 \times 10^{-3}$. 
Please refer to ~\cref{sec:method} for definitions of the above hyperparameters.

\subsubsection*{A.2. Hierarchical Sampling}
\label{sec:supp-hier-sampling}

We first organize the segmentation masks into a hierarchical structure determined by the inclusion ratio between them. One mask $A$ is designated as a \textit{child} of another mask $B$ when$\frac{|A\cap B|}{|A|} > p_{in}$ and $\frac{|A\cap B|}{|A\cup B|} < p_{IoU}$. We empirically set $p_{in} = 0.95$ and $p_{IoU} = 0.85$. 
We present the hierarchical sampling algorithm we introduced in \cref{sec:hier-sampling} in \cref{alg:data_sampling}.

\begin{algorithm}[h]
\caption{Data Sampling}
\label{alg:data_sampling}
\begin{algorithmic}
    \STATE pos\_samples $\gets$ []
    \STATE neg\_samples $\gets$ []
    \FORALL{$A$ $\in$ leaf\_masks}
        \STATE  \# Sampling positive pairs from the leaf node
        \STATE sample $\gets$ (Random($A$), Random($A$)) 
        \WHILE{$A$ has Parent}
            \STATE $B$ $\gets$ $A$.Parent
            \STATE pos\_samples += sample
            \STATE \# Sampling negative pairs for the current level
            \STATE sample $\gets$ (Random($A$), Random($\bar{A} \cap B$))
            \STATE neg\_samples += sample
            \STATE $A$ $\gets$ $B$            
        \ENDWHILE 
    \ENDFOR
    \RETURN pos\_samples, neg\_samples
    
\end{algorithmic}
\end{algorithm}

We sample same number of positive pairs and negative pairs for training for training efficiency. Implementing a 4-1 ratio ($4\times$ more negatives than positives) instead of 1-1, the normalized covering (NC) score increases from 0.709 to \textbf{0.720}. However, this slow down our training by 43\%, primarily due to the time-consuming computation of ultrametric distances and the associated minimum spanning tree.

\subsubsection*{A.3. Additional Details on Evaluation Metrics}
\label{sec:imple-metrics}
\noindent\paragraph{Normalized Covering Score}

As discribed in \cref{sec:dataset}, we measure the \textit{quality} of hierarchical segmentation with the Normalized Covering (NC) score~\cite{ke2022unsupervised}. This metric averages the Intersection over Union (IoU) between each ground truth mask and the \textit{best-matching} (\ie most-overlapping) predicted mask. 
The metric is defined as \begin{equation}
     \mathrm {NC}(S' \rightarrow  S) = \frac {1}{|S|} \sum \limits _{A \in S} \max \limits _{A' \in S'} \frac {|A \cap A'|}{|A \cup A'|}
\end{equation}
Where $S$ denotes all segmentation masks and $S'$ denotes all predicted masks. For LSeg, DFF, and LeRF, which only output feature fields without segmentation results ($S'$), we adopt a similar approach as our method, and we extract segmentations by thresholding feature distances.

\noindent\paragraph{Segmentation Injectivity Score}
We propose the Segmentation Injectivity (SI) score to measure if each pixel belongs to only one mask for each level of granularity. As described in \cref{sec:dataset}, given a ground truth mask, we first randomly sample $p_1$ and $p_2$ from that mask, and then query the model at these points and granularity for a new mask prediction. Then, we measure the IoU between the two resulting masks. We iterate this process $N=100$ for each ground truth mask, calculating scores for each run. The final SI score is obtained by averaging the scores across all ground truth masks and viewpoints.

We represent the segmentation model as $F(v, p, t) \rightarrow A'$ where $v$ denotes the viewpoint, $p$ represents the pixel query, $t$ corresponds to the granularity level, and $A'$ is the resulting segmentation mask. The SI score is defined as \begin{align*}
  \mathrm {SI}(S' \rightarrow  S) &= \frac {1}{|N||S|} \sum \limits _{A \in S} \sum^N \limits _{i=1} \frac {|F(v, p_1^i, t) \cap F(v, p_2^i, t)|}{|(F(v, p_1^i, t) \cup F(v, p_2^i, t)|} \\
     \quad \text{where} \quad & t = \arg\max_{t} \frac{|A \cap |F(v, p_1^i, t)|}{|A \cup |F(v, p_1^i, t)|}  
\end{align*}
where $v$ represents the view corresponding to the ground truth mask $A$.

\begin{figure}[t]
    
    \centering
    \begin{subfigure}[t]{0.33\linewidth}
        \centering
        \includegraphics[height=1.0in]{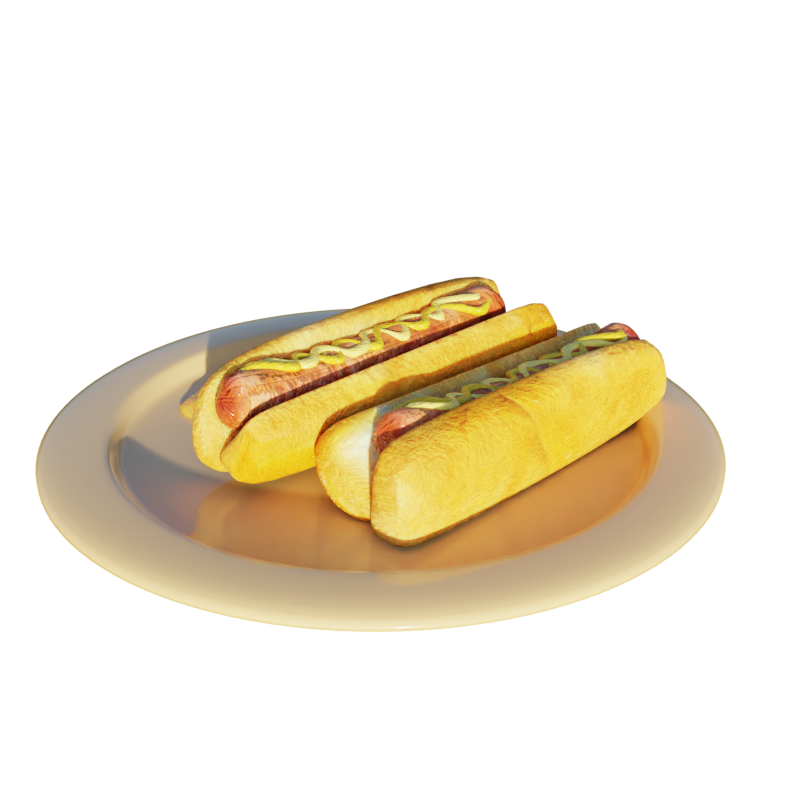}
        \caption{Image}
    \end{subfigure}%
    ~ 
    \begin{subfigure}[t]{0.33\linewidth}
        \centering
        \includegraphics[height=1.0in]{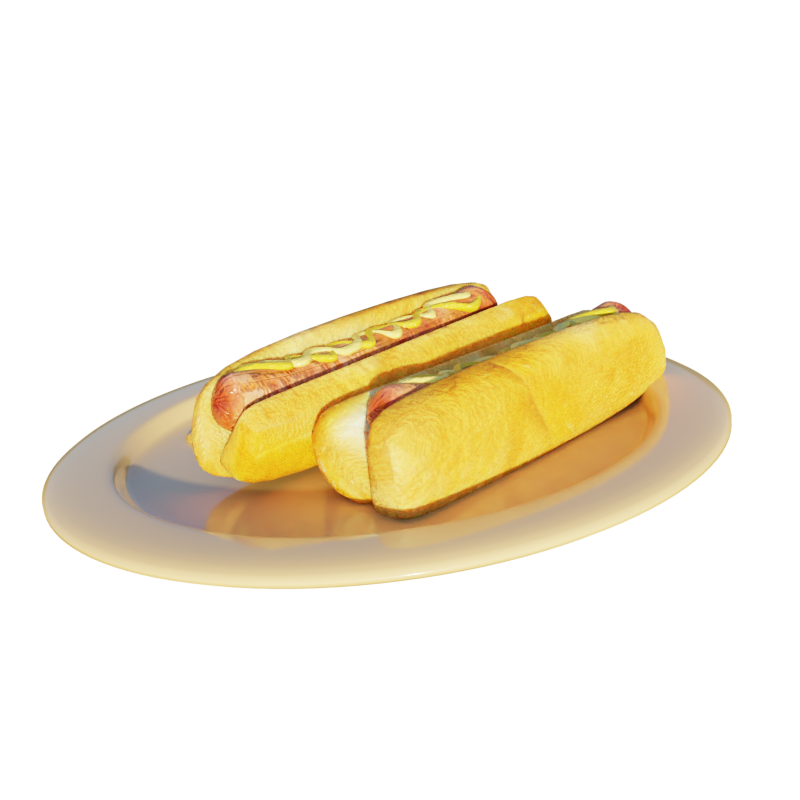}
        \caption{Shifted Image}
    \end{subfigure}%
    ~ 
    \begin{subfigure}[t]{0.33\linewidth}
        \centering
        \includegraphics[height=1.0in]{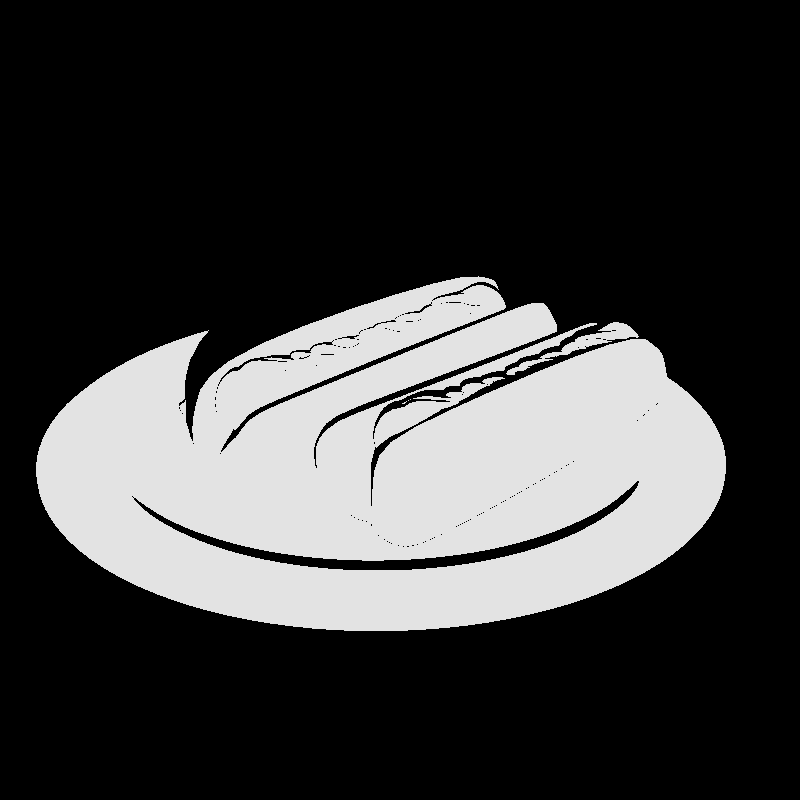}
        \caption{Visibility Mask}
    \end{subfigure}
    \caption{\textbf{View Consistency}: We evaluate the view consistency between the source viewpoint (a) and another one shifted 10 degrees (b). We render the ground truth visibility mask (c) with ray casting to avoid the occlusion/disocclusion between views.}
    \label{fig:view-consistency}
\end{figure}

\noindent\paragraph{View Consistency Score}
We use the View Consistency (VC) score to measure the 3D consistency of image segmentations. Starting with the source view, we rotate the camera by 10 degrees, rendering both a new image and the corresponding ground truth visibility mask in the shifted view -- \cref{fig:view-consistency} provides an example of two viewpoints and their visibility mask on the Blender Hotdog scene.

For a given point query $p_1$ and a granularity $t$ and its mask prediction $A_1 = F(v, p_1, t)$ in the source view, we leverage the ground truth camera parameters to warp the point to $p_2 = T(p_1)$ and the mask prediction to $T(A_1)$ in the shifted view where $T$ denotes the pixel transformation. Following this, we query the model in the shifted view with $p_2$ using the same threshold $t$, resulting in $A_2 = F(v', p_2, t)$.

Utilizing the visibility mask $V$, we eliminate pixels that are occluded in either view from $A_2$ and $T(A_1)$. The Intersection over Union (IoU) between the remaining masks is computed as the VC score for this sample. We reduce the noise induced by random sampling by computing this score for $N = 100$ times per ground truth mask.

Taken together, the VC score is defined as \begin{align*}
  \mathrm {VC}(S' \rightarrow  S) &= \frac {1}{|N||S|} \sum \limits _{A \in S} \sum^N \limits _{i=1} \frac {|T(A_1^i) \cap A_2^i \cap V|}{|(T(A_1^i) \cup A_2^i) \cap V|} \\
     \quad \text{where} \quad & A_1^i = F(v, p_1^i, \arg\max_{t} \frac{|A \cap |F(v, p_1^i, t)|}{|A \cup |F(v, p_1^i, t)|} )\\
     & A_2^i = F(v', p_2^i, \arg\max_{t} \frac{|A \cap |F(v, p_1^i, t)|}{|A \cup |F(v, p_1^i, t)|} )\\
\end{align*}
For additional details, please refer to \cref{sec:dataset}.

We also evaluate the View Consistency across multiple angles, in \cref{tab:vc_angles}. The ranking of the methods is the same.

\setlength{\tabcolsep}{6pt}
\begin{table}[h]
  \caption{View Consistency score with different view angles.
  }
  \vspace{-0.5mm}
  \label{tab:vc_angles}
  \centering
  \small
  \resizebox{0.78\linewidth}{!}{
  \begin{tabular}{@{}lcccc@{}}
    \toprule
    %Heading level & Example & Font size and style\\
    
    Method & $\text{VC}_{10^\circ}$ $\uparrow$ & $\text{VC}_{45^\circ}$ $\uparrow$ & $\text{VC}_{90^\circ}$ $\uparrow$ & $\text{VC}_{135^\circ}$ $\uparrow$\\[-0.8mm] %\cline{2-5} 
    \midrule
   LSeg [26] &0.536 & 0.522 & 0.510 & 0.498\\[-0.3mm]
   LSeg + DFF [24] & \textbf{0.827} & \textbf{0.813}&\textbf{0.810}&\textbf{0.808}\\[-0.3mm]
   SAM3D [6] & 0.724 & 0.601 & 0.578 & 0.539\\
   \hline
   Ours & 0.789 & 0.763 & 0.742 & 0.712\\[-0.8mm]
  \bottomrule
  \end{tabular}
  }
\end{table}

\noindent\paragraph{Depth Error}
We leverage the ground truth depth map rendered in blender to compute the depth error of our method. The scale of the depth error adheres to the normalized NeRF scene.

\begin{figure}[b]
    \hspace{-3mm}
    % \begin{center}
    \centering
    \includegraphics[width=0.6\linewidth]
    % \end{center}
    {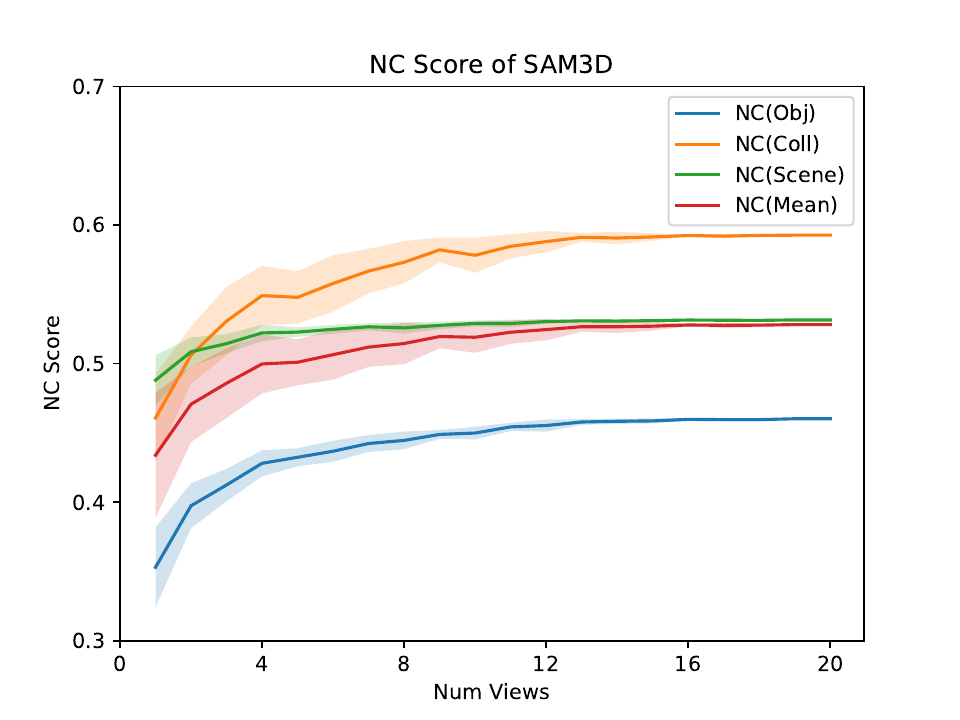}
        \caption{\textbf{NC Score of SAM3D}: We present the Normalized Covering (NC) score (y-axis) of SAM3D, correlating it with the number of views (x-axis) from which we propagate the SAM segmentation masks.}
    \label{fig:sam3d}
\end{figure}

\begin{table}[tb]
  \caption{\textbf{Distilled Feature Fields}: 
    We present the NC score of DFF with volume rendering (VR) and the NC score of the official codebase on our BlenderHS dataset.}
  \label{tab:dff}
  \centering
  \begin{tabular}{@{}lcccc@{}}
    \toprule
    Method & NC$_{\text{obj}}$ $\uparrow$ & NC$_{\text{coll}}$ $\uparrow$ & NC$_{\text{scene}}$ $\uparrow$ & NC$_{\text{mean}}$ $\uparrow$ \\ %\cline{2-5} 
    \midrule
           DFF (VR) & \textbf{0.164} &	\textbf{0.427}&	\textbf{0.839}	& \textbf{0.477}\\
           DFF (Official) & 0.082 &	0.286&	0.666	&0.345\\
    \bottomrule
  \end{tabular}
\end{table}

\subsubsection{A.4. Additional Details on Baselines}
\label{sec:imple-models}
\noindent\paragraph{DFF}

We configure DFF to use a white background and employ uniform ray sampling on the BlenderHS dataset. All other hyperparameters directly adhere to the official implementation.
Furthermore, DFF's official code\footnote{\url{https://github.com/pfnet-research/distilled-feature-fields}} does not apply volume rendering to the feature branch. Instead, it generates a feature map by directly querying the volumetric features at the 3D locations of the predicted surface points.
We extended their code to perform volume rendering, and we show that using volume rendering leads to improved performance on the BlenderHS dataset (see \cref{tab:dff}).

\noindent\paragraph{LeRF}
We use LeRF's reported NSVF hyperparameters for the Blender synthetic dataset. This includes configuring the background to white, selecting uniform sampling as the ray sampling strategy, disabling space distortion, and setting average appearance embedding to off. We train the model for 20000 steps.
For the Normalized Covering Score, we report the highest result among all 30 semantic scales available in the LeRF feature field for each ground truth granularity. For the Segmentation Injectivity score and View Consistency scores, we evaluate LeRF at the semantic scale corresponding to the ground truth granularity which yields the highest NC score.

\noindent\paragraph{SAM3D}

Given a pretrained NeRF and a segmentation mask from a single view, SAM3D optimizes a binary voxel grid using mask inverse rendering and cross-view self-prompting to propagate the mask into 3D. In our experiments, we propagate the SAM masks from the segment-everything mode using 20 training views (while still using all 100 training images to pretrain the NeRF). We observed saturation in SAM3D's NC score after 20 views, and, on an A6000 GPU, it takes approximately a day per scene to propagate the segmentation maps from 20 views. In contrast, our method takes around 2 hours.

\noindent\paragraph{SAM}

We employ the ViT-H model from the official SAM GitHub repository\footnote{\url{https://github.com/facebookresearch/segment-anything}} to generate mask predictions. To generate the training data of our model, we use the segment-everything mode to generate our supervision.

In the evaluation process, when querying segmentation models with a randomly sampled point, we employ the point as a prompt for SAM to generate the segmentation prediction. This approach, compared to evaluating based on the output of the segment-everything mode, yields a higher NC score and provides a clearer granularity level.

\subsubsection{A.5. Training and Inference Time}
We train and perform inference on a Titan RTX GPU. Training typically takes $\sim$70 minutes, while inference takes 5 seconds per granularity for 10 views. The main expense in inference is the watershed algorithm running on 3D point clouds, which is executed once per granularity and is view-independent.

\subsection* {B. Additional Qualitative Results}
\label{sec:qualitative}
\subsubsection{B.1. BlenderHS Dataset}

\begin{figure}[t]
    \centering
    \begin{subfigure}[t]{0.23\linewidth}
        \centering
        \includegraphics[height=1.0in]{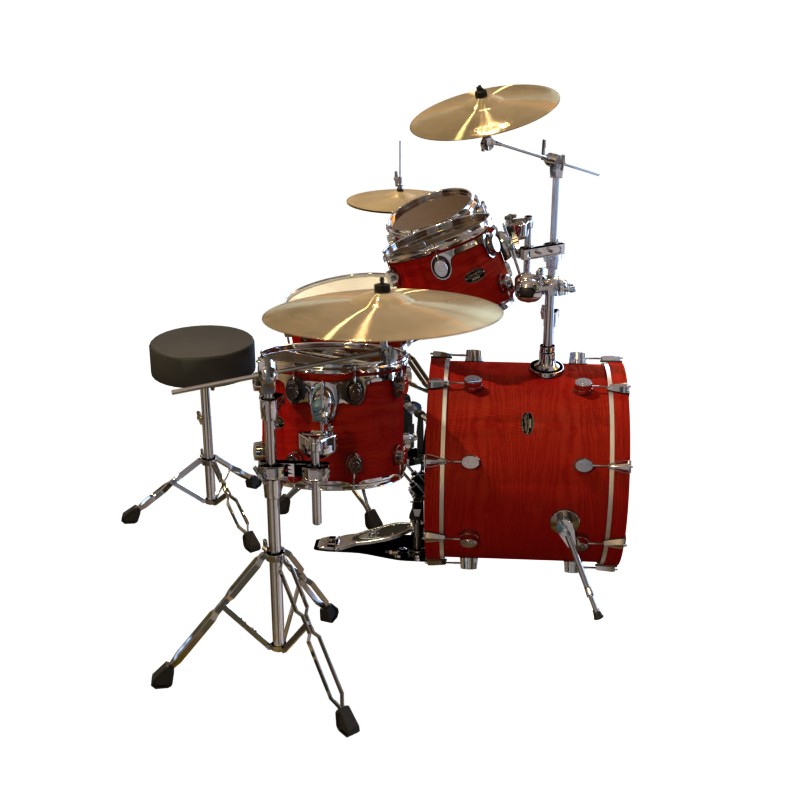}
    \end{subfigure}%
    ~ 
    \begin{subfigure}[t]{0.23\linewidth}
        \centering
        \includegraphics[height=1.0in]{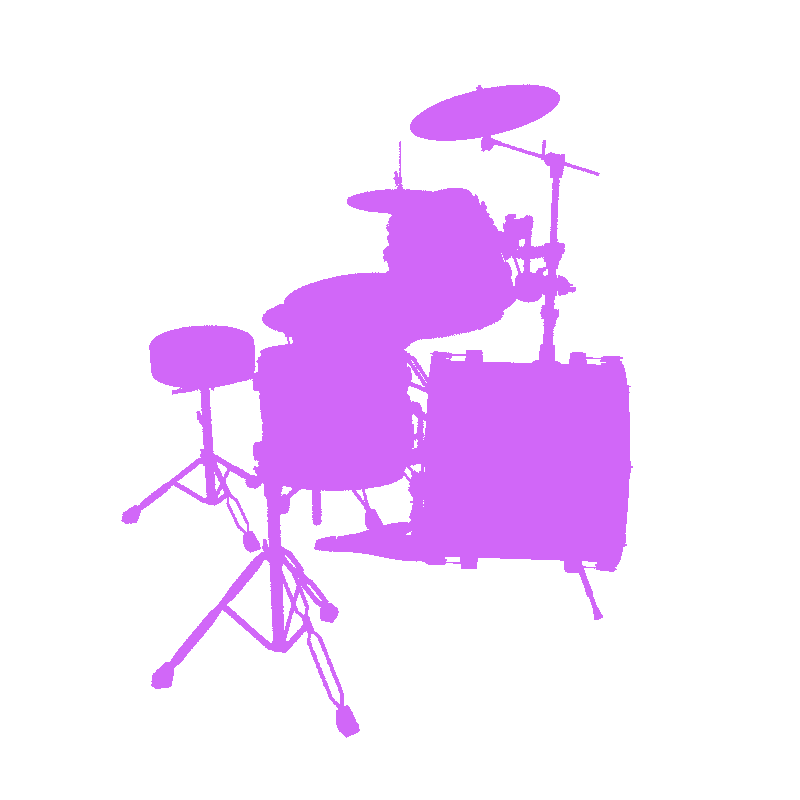}
    \end{subfigure}%
    ~ 
    \begin{subfigure}[t]{0.23\linewidth}
        \centering
        \includegraphics[height=1.0in]{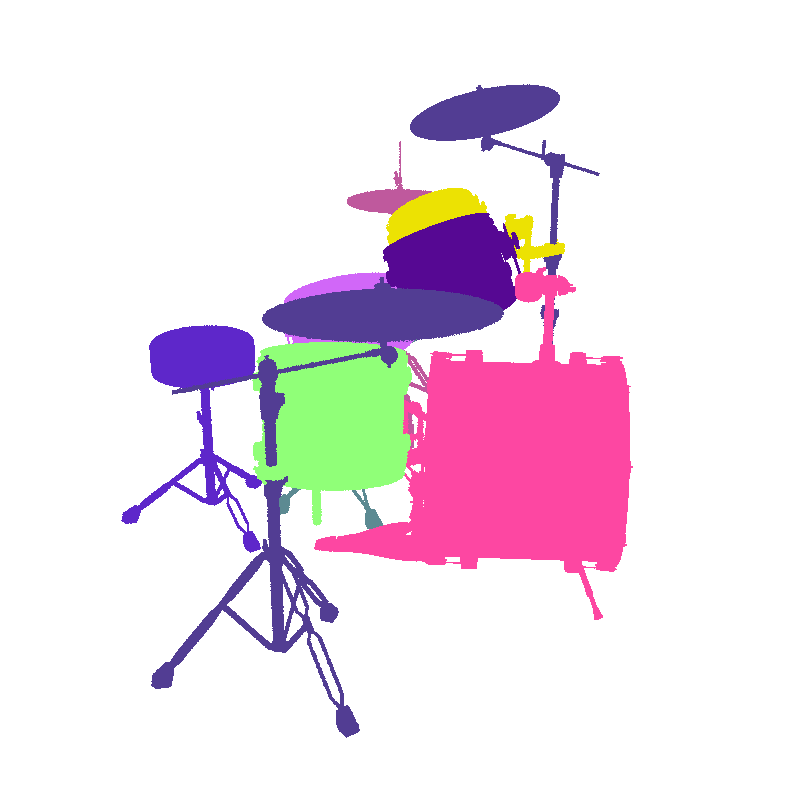}
    \end{subfigure}
    \begin{subfigure}[t]{0.23\linewidth}
        \centering
        \includegraphics[height=1.0in]{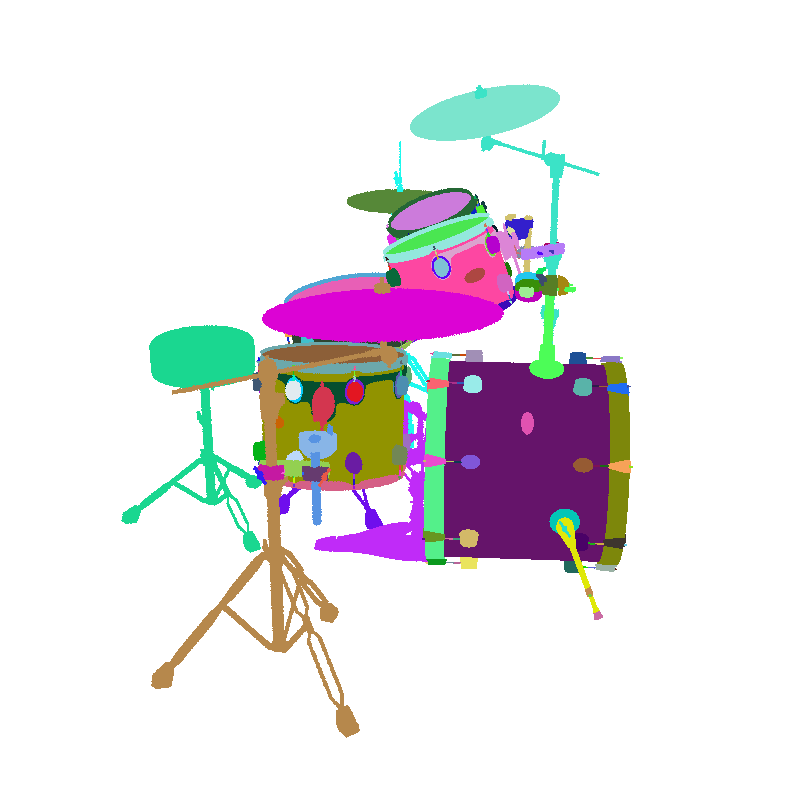}
    \end{subfigure}
    \caption{\textbf{Blender with Hierarchical Segmentation (Blender-HS)}: We render hierarchical segmentation maps at three levels of granularity, namely \textit{Scene}, \textit{Collection}, and \textit{Object}, using information saved into the blender file by the scene artist.} % for the synthetic dataset with the blender file. }
    \label{fig:seg_maps}

\end{figure}

\begin{figure*}[t]
    \centering
    \begin{subfigure}[t]{0.25\linewidth}
        \centering
        \includegraphics[height=1.1in]{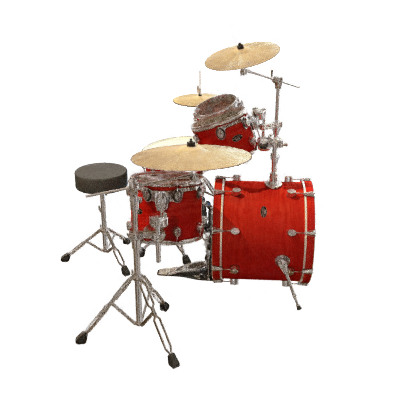}
    \end{subfigure}%
    ~ 
    \begin{subfigure}[t]{0.25\linewidth}
        \centering
        \includegraphics[height=1.1in]{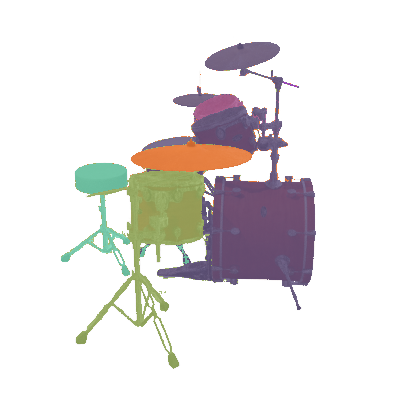}
    \end{subfigure}%
    ~ 
    \begin{subfigure}[t]{0.25\linewidth}
        \centering
        \includegraphics[height=1.1in]{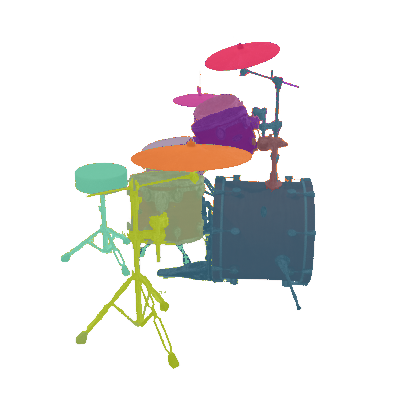}
    \end{subfigure}%
    ~ 
    \begin{subfigure}[t]{0.25\linewidth}
        \centering
        \includegraphics[height=1.1in]{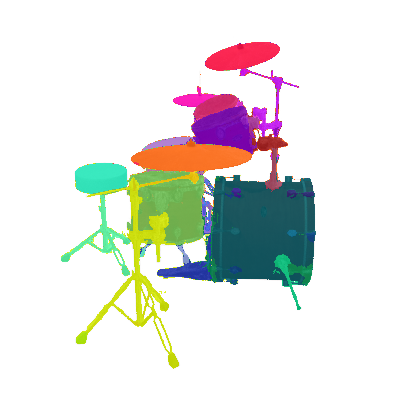}
    \end{subfigure}
    \medskip
    \vspace{-5mm}
    \begin{subfigure}[t]{0.25\linewidth}
        \centering
        \includegraphics[height=1.1in]{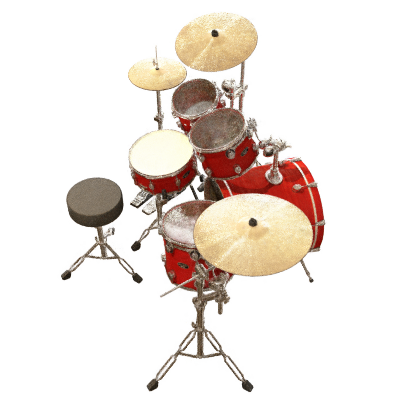}
    \end{subfigure}%
    ~ 
    \begin{subfigure}[t]{0.25\linewidth}
        \centering
        \includegraphics[height=1.1in]{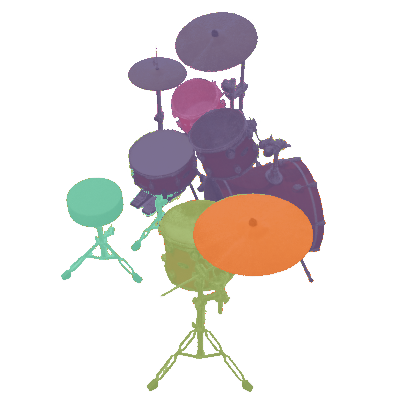}
    \end{subfigure}%
    ~ 
    \begin{subfigure}[t]{0.25\linewidth}
        \centering
        \includegraphics[height=1.1in]{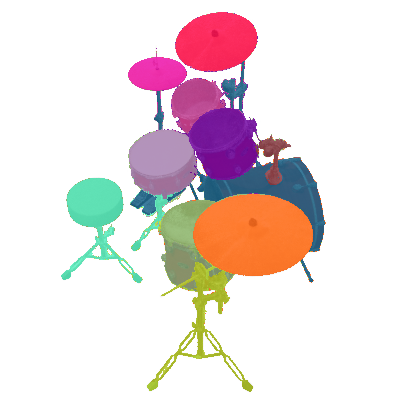}
    \end{subfigure}%
    ~ 
    \begin{subfigure}[t]{0.25\linewidth}
        \centering
        \includegraphics[height=1.1in]{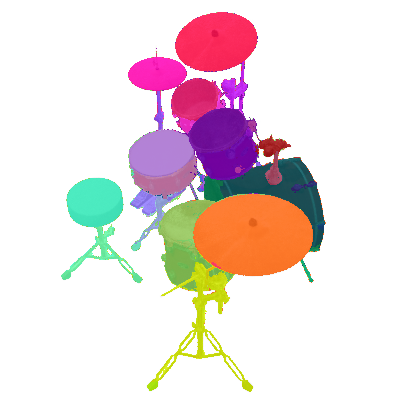}
    \end{subfigure}
    \medskip
    \vspace{2mm}
    \begin{subfigure}[t]{0.25\linewidth}
        \centering
        \includegraphics[height=1.1in]{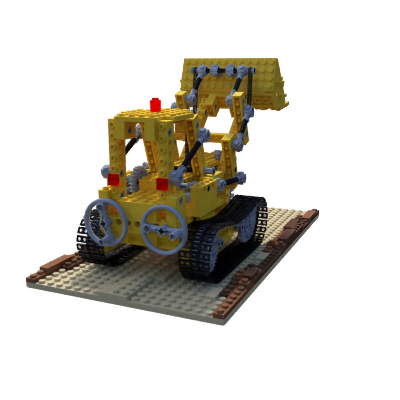}
    \end{subfigure}%
    ~ 
    \begin{subfigure}[t]{0.25\linewidth}
        \centering
        \includegraphics[height=1.1in]{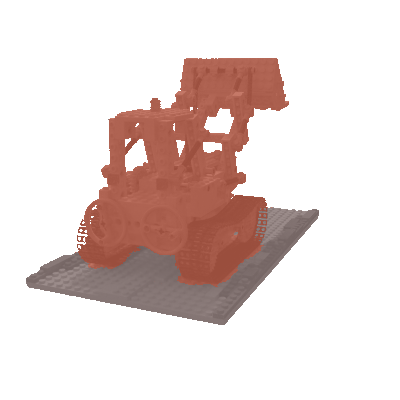}
    \end{subfigure}%
    ~ 
    \begin{subfigure}[t]{0.25\linewidth}
        \centering
        \includegraphics[height=1.1in]{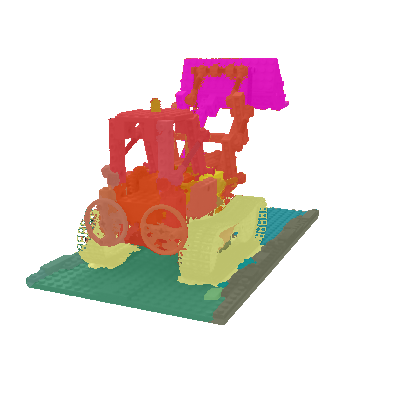}
    \end{subfigure}%
    ~ 
    \begin{subfigure}[t]{0.25\linewidth}
        \centering
        \includegraphics[height=1.1in]{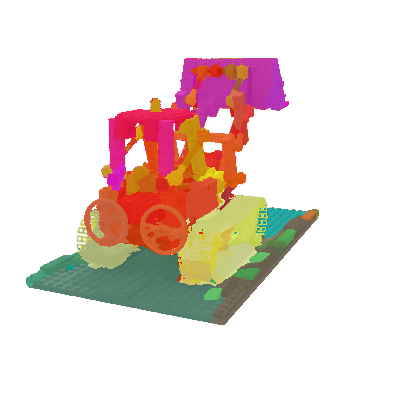}
    \end{subfigure}
    
    \medskip
    \vspace{-8mm}
    \begin{subfigure}[t]{0.25\linewidth}
        \centering
        \includegraphics[height=1.1in]{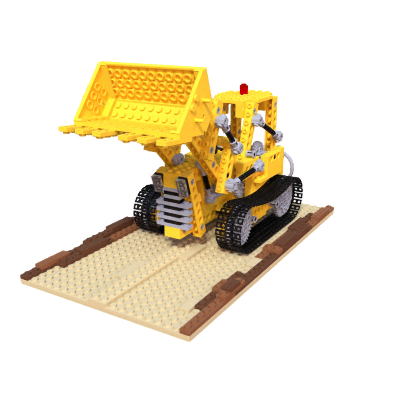}
    \end{subfigure}%
    ~ 
    \begin{subfigure}[t]{0.25\linewidth}
        \centering
        \includegraphics[height=1.1in]{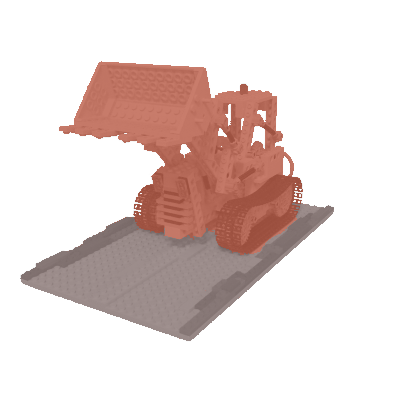}
    \end{subfigure}%
    ~ 
    \begin{subfigure}[t]{0.25\linewidth}
        \centering
        \includegraphics[height=1.1in]{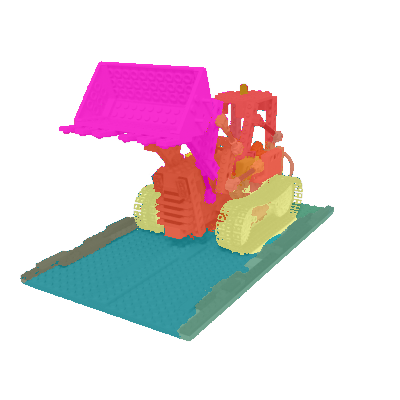}
    \end{subfigure}%
    ~ 
    \begin{subfigure}[t]{0.25\linewidth}
        \centering
        \includegraphics[height=1.1in]{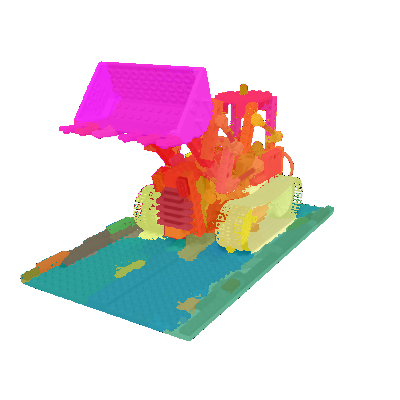}
    \end{subfigure}
    
    \medskip
    \vspace{-5mm}
    \begin{subfigure}[t]{0.25\linewidth}
        \centering
        \includegraphics[height=1.1in]{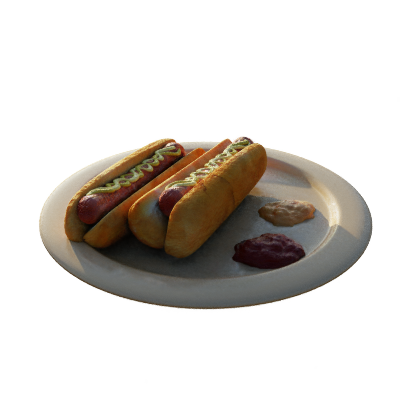}
    \end{subfigure}%
    ~ 
    \begin{subfigure}[t]{0.25\linewidth}
        \centering
        \includegraphics[height=1.1in]{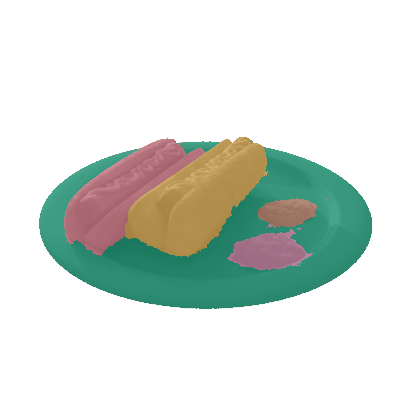}
    \end{subfigure}%
    ~ 
    \begin{subfigure}[t]{0.25\linewidth}
        \centering
        \includegraphics[height=1.1in]{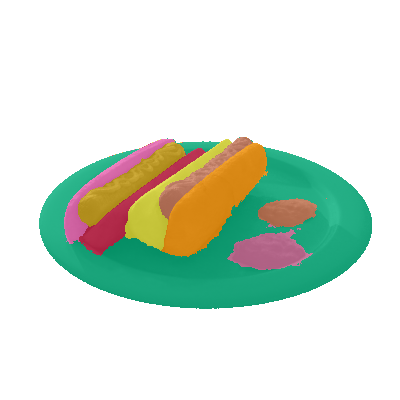}
    \end{subfigure}%
    ~ 
    \begin{subfigure}[t]{0.25\linewidth}
        \centering
        \includegraphics[height=1.1in]{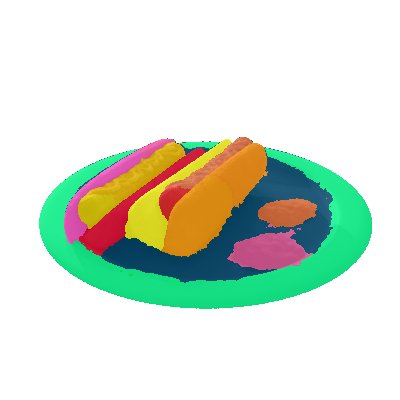}
    \end{subfigure}
    
    \medskip
    \vspace{-8mm}
    \begin{subfigure}[t]{0.25\linewidth}
        \centering
        \includegraphics[height=1.1in]{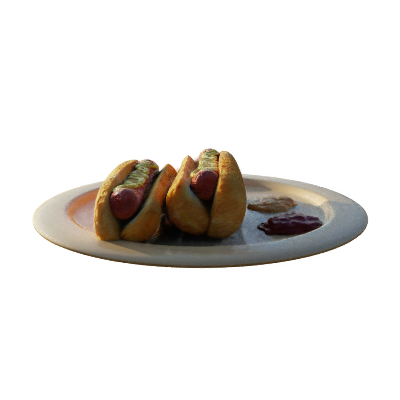}
        \caption{Image}
    \end{subfigure}%
    ~ 
    \begin{subfigure}[t]{0.25\linewidth}
        \centering
        \includegraphics[height=1.1in]{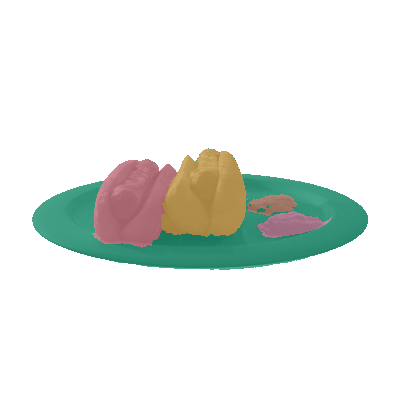}
    \end{subfigure}%
    ~ 
    \begin{subfigure}[t]{0.25\linewidth}
        \centering
        \includegraphics[height=1.1in]{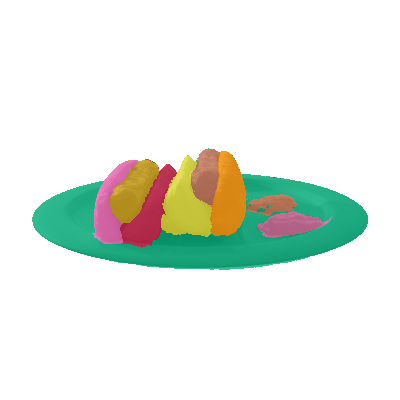}
        \caption{Hierarchical Segmentation}
    \end{subfigure}%
    ~ 
    \begin{subfigure}[t]{0.25\linewidth}
        \centering
        \includegraphics[height=1.1in]{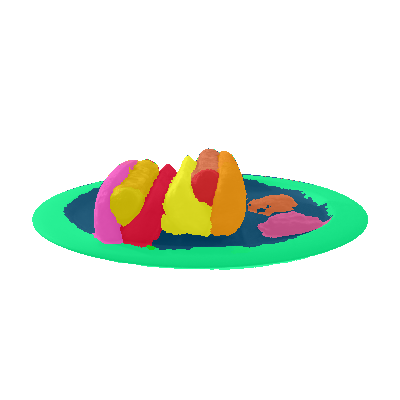}
    \end{subfigure}
    \caption{\textbf{BlenderHS Dataset}: We present the qualitative results obtained from our BlenderHS dataset. The segmentation results demonstrate both view consistency and hierarchical structure.}
    \label{fig:blender-hs}
\end{figure*}

\begin{figure}[t]
    \centering
    \begin{subfigure}[t]{0.33\linewidth}
        \centering
        \includegraphics[height=1.0in]{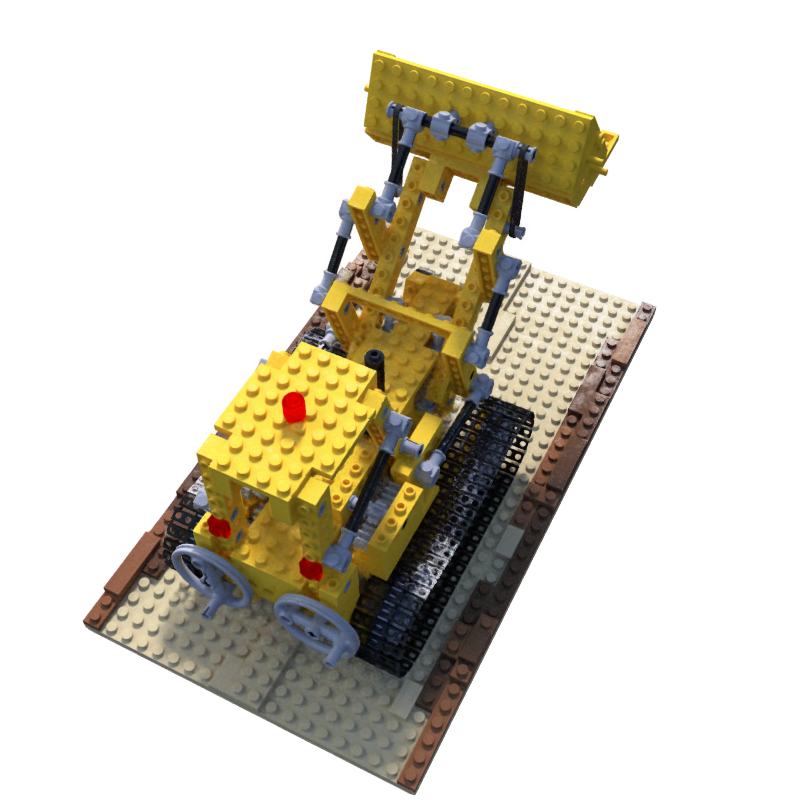}
        \caption{Image}
    \end{subfigure}%
    ~ 
    \begin{subfigure}[t]{0.33\linewidth}
        \centering
        \includegraphics[height=1.0in]{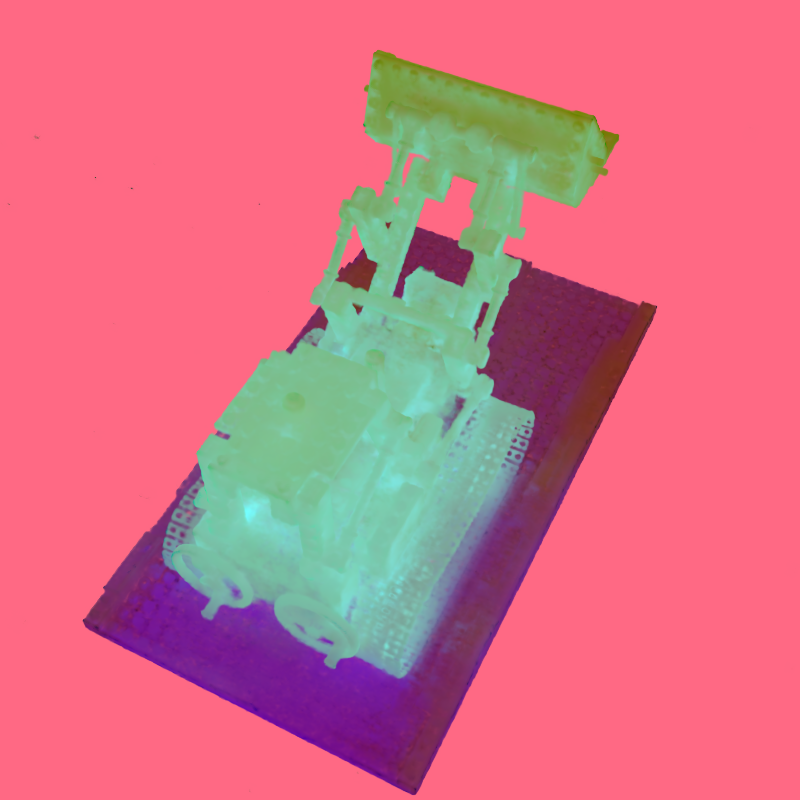}
        \caption{DFF~\cite{kobayashi2022distilledfeaturefields}}
    \end{subfigure}%
    ~ 
    \begin{subfigure}[t]{0.33\linewidth}
        \centering
        \includegraphics[height=1.0in]{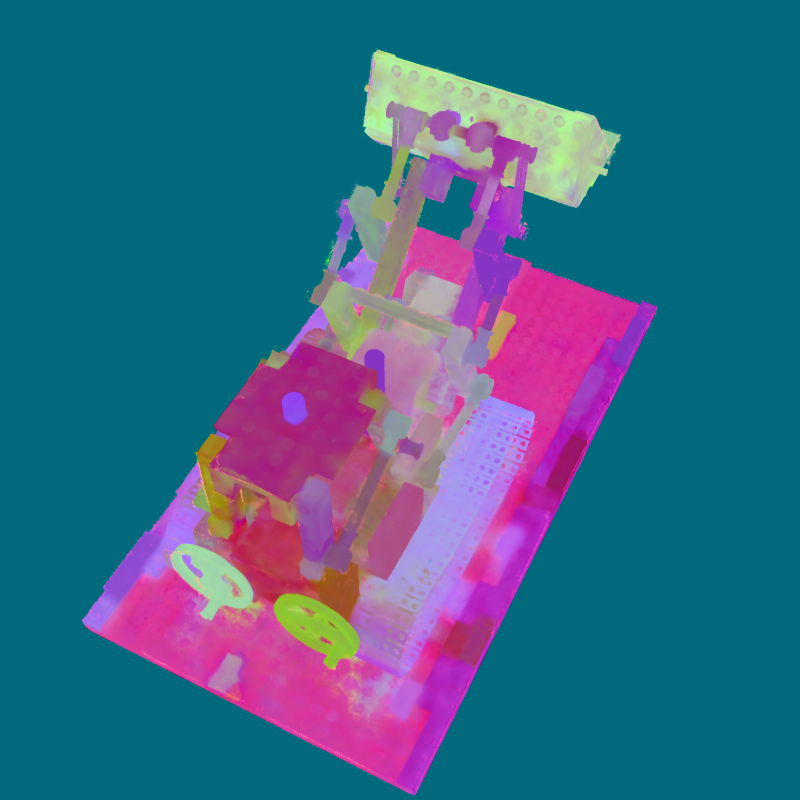}
        \caption{Ours}
    \end{subfigure}
    \caption{\textbf{Feature Visualization}: We visualize rendered feature maps using PCA. The feature map generated by DFF~\cite{kobayashi2022distilledfeaturefields} fails to distinguish between different parts of the Lego. In contrast, our method learns features that can distinguish various Lego bricks.}
    \label{fig:feat-fields}
\end{figure}

We first visualize the ground truth segmentations for the Drums scene in the BlenderHS Dataset~\cite{mildenhall2020nerf} in \cref{fig:seg_maps}. We then present qualitative results on the BlenderHS dataset~\cite{mildenhall2020nerf} in \cref{fig:blender-hs}. Our segmentations exhibit a hierarchical structure and maintain consistency across different views. We also visualize our ultrametric feature field on the Lego scene in \cref{fig:feat-fields}, showing sharper features than DFF.

\subsubsection{B.2. PartNet Dataset}
\begin{figure*}[t]
    \centering
    \begin{subfigure}[t]{0.25\linewidth}
        \centering
        \includegraphics[height=1.1in]{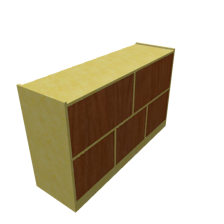}
    \end{subfigure}%
    ~ 
    \begin{subfigure}[t]{0.25\linewidth}
        \centering
        \includegraphics[height=1.1in]{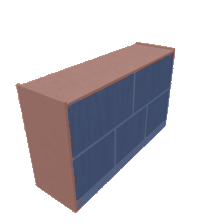}
    \end{subfigure}%
    ~ 
    \begin{subfigure}[t]{0.25\linewidth}
        \centering
        \includegraphics[height=1.1in]{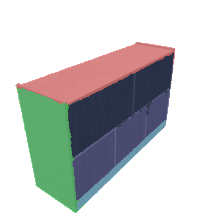}
    \end{subfigure}%
    ~ 
    \begin{subfigure}[t]{0.25\linewidth}
        \centering
        \includegraphics[height=1.1in]{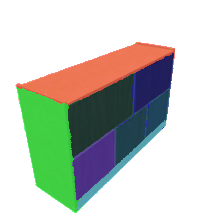}
    \end{subfigure}
    \medskip
    \vspace{-8mm}
    \begin{subfigure}[t]{0.25\linewidth}
        \centering
        \includegraphics[height=1.1in]{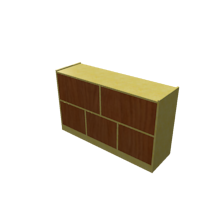}
    \end{subfigure}%
    ~ 
    \begin{subfigure}[t]{0.25\linewidth}
        \centering
        \includegraphics[height=1.1in]{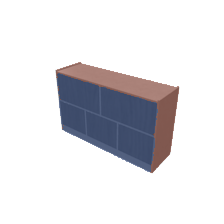}
    \end{subfigure}%
    ~ 
    \begin{subfigure}[t]{0.25\linewidth}
        \centering
        \includegraphics[height=1.1in]{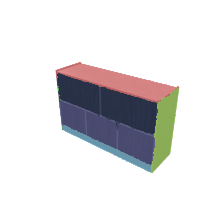}
    \end{subfigure}%
    ~ 
    \begin{subfigure}[t]{0.25\linewidth}
        \centering
        \includegraphics[height=1.1in]{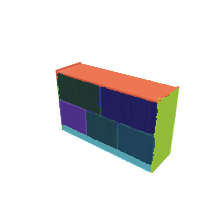}
    \end{subfigure}
    \medskip
    \vspace{-7mm}
    \begin{subfigure}[t]{0.25\linewidth}
        \centering
        \includegraphics[height=1.1in]{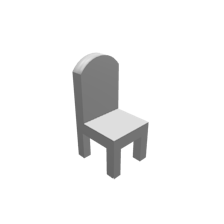}
    \end{subfigure}%
    ~ 
    \begin{subfigure}[t]{0.25\linewidth}
        \centering
        \includegraphics[height=1.1in]{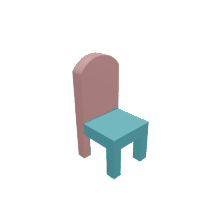}
    \end{subfigure}%
    ~ 
    \begin{subfigure}[t]{0.25\linewidth}
        \centering
        \includegraphics[height=1.1in]{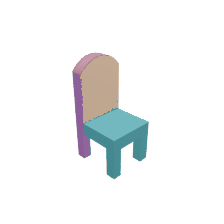}
    \end{subfigure}%
    ~ 
    \begin{subfigure}[t]{0.25\linewidth}
        \centering
        \includegraphics[height=1.1in]{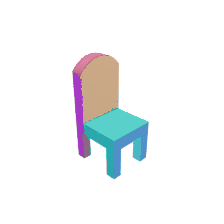}
    \end{subfigure}
    
    \medskip
    \vspace{-8mm}
    \begin{subfigure}[t]{0.25\linewidth}
        \centering
        \includegraphics[height=1.1in]{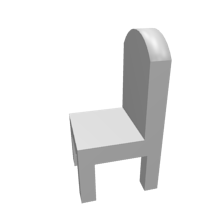}
    \end{subfigure}%
    ~ 
    \begin{subfigure}[t]{0.25\linewidth}
        \centering
        \includegraphics[height=1.1in]{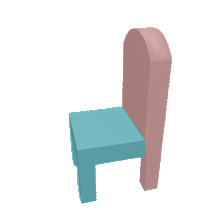}
    \end{subfigure}%
    ~ 
    \begin{subfigure}[t]{0.25\linewidth}
        \centering
        \includegraphics[height=1.1in]{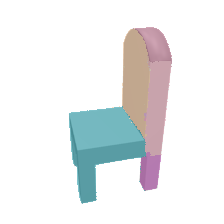}
    \end{subfigure}%
    ~ 
    \begin{subfigure}[t]{0.25\linewidth}
        \centering
        \includegraphics[height=1.1in]{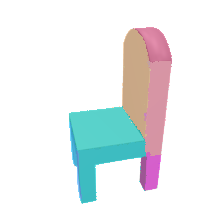}
    \end{subfigure}
    
    \medskip
    \vspace{0mm}
    \begin{subfigure}[t]{0.25\linewidth}
        \centering
        \includegraphics[height=1.1in]{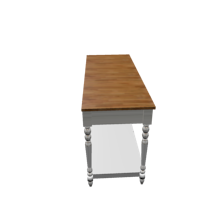}
    \end{subfigure}%
    ~ 
    \begin{subfigure}[t]{0.25\linewidth}
        \centering
        \includegraphics[height=1.1in]{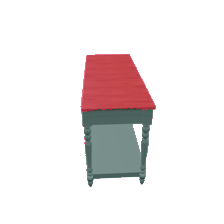}
    \end{subfigure}%
    ~ 
    \begin{subfigure}[t]{0.25\linewidth}
        \centering
        \includegraphics[height=1.1in]{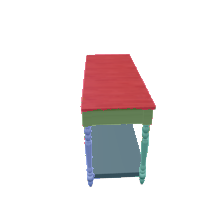}
    \end{subfigure}%
    ~ 
    \begin{subfigure}[t]{0.25\linewidth}
        \centering
        \includegraphics[height=1.1in]{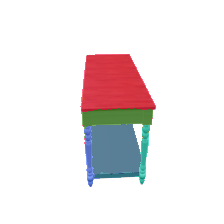}
    \end{subfigure}
    
    \medskip
    \vspace{-8mm}
    \begin{subfigure}[t]{0.25\linewidth}
        \centering
        \includegraphics[height=1.1in]{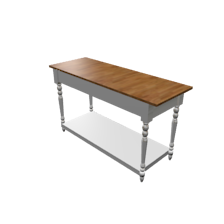}
        \caption{Image}
    \end{subfigure}%
    ~ 
    \begin{subfigure}[t]{0.25\linewidth}
        \centering
        \includegraphics[height=1.1in]{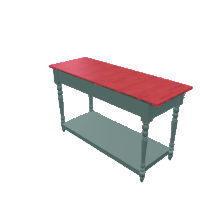}
    \end{subfigure}%
    ~ 
    \begin{subfigure}[t]{0.25\linewidth}
        \centering
        \includegraphics[height=1.1in]{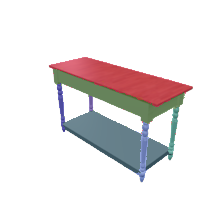}
        \caption{Hierarchical Segmentation}
    \end{subfigure}%
    ~ 
    \begin{subfigure}[t]{0.25\linewidth}
        \centering
        \includegraphics[height=1.1in]{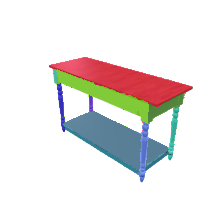}
    \end{subfigure}
    \caption{\textbf{PartNet Dataset}: We showcase the qualitative results on the PartNet dataset.}
    \label{fig:partnet}
\end{figure*}

We present qualitative results on the PartNet dataset~\cite{Mo_2019_CVPR} in \cref{fig:partnet}. Our method is able to generate hierarchical segmentation results of different objects. Leveraging the 2D masks predicted with SAM as guidance, our method proficiently segments various surfaces of sub-parts within the object, while those are not included in the PartNet ground truth annotations.

\subsubsection{B.3. LLFF Dataset}

\begin{figure*}[t]
    \centering
    \begin{subfigure}[t]{0.25\linewidth}
        \centering
        \includegraphics[height=0.8in]{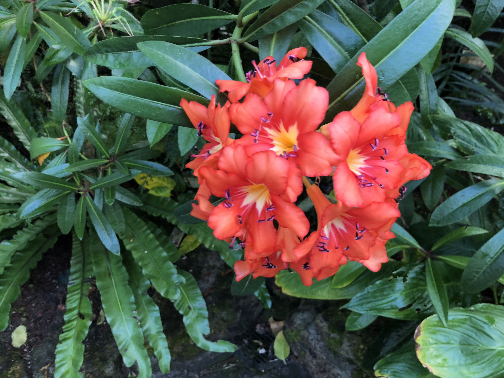}
    \end{subfigure}%
    ~ 
    \begin{subfigure}[t]{0.25\linewidth}
        \centering
        \includegraphics[height=0.8in]{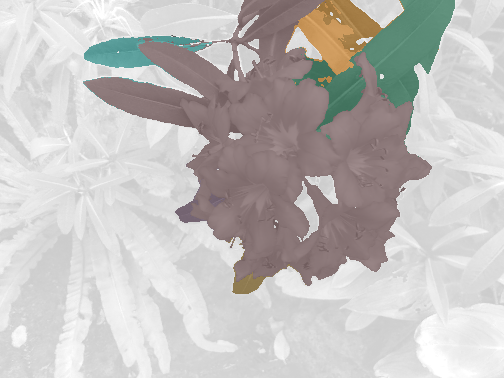}
    \end{subfigure}%
    ~ 
    \begin{subfigure}[t]{0.25\linewidth}
        \centering
        \includegraphics[height=0.8in]{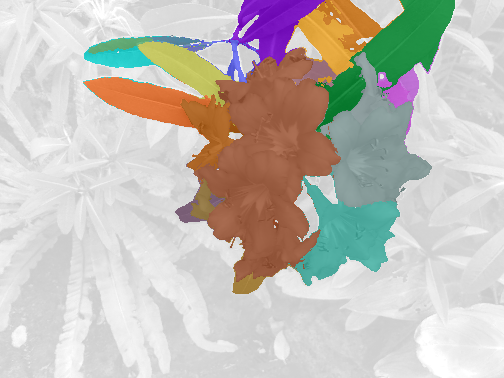}
    \end{subfigure}%
    ~ 
    \begin{subfigure}[t]{0.25\linewidth}
        \centering
        \includegraphics[height=0.8in]{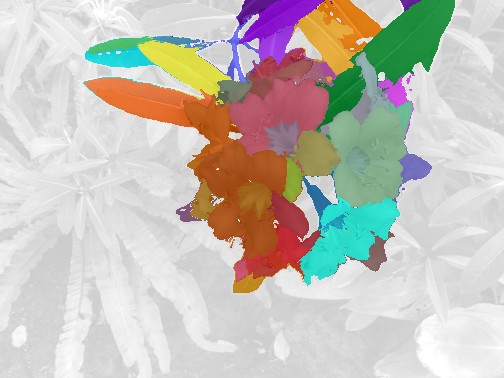}
    \end{subfigure}
    \medskip
    \vspace{-5mm}
    \begin{subfigure}[t]{0.25\linewidth}
        \centering
        \includegraphics[height=0.8in]{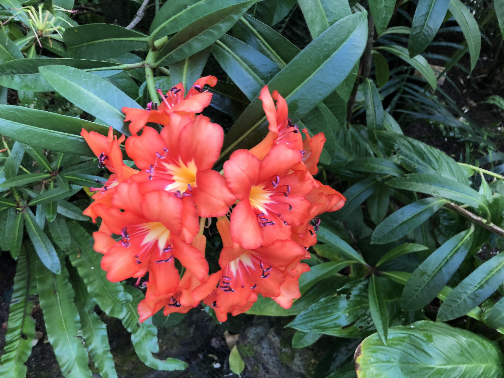}
    \end{subfigure}%
    ~ 
    \begin{subfigure}[t]{0.25\linewidth}
        \centering
        \includegraphics[height=0.8in]{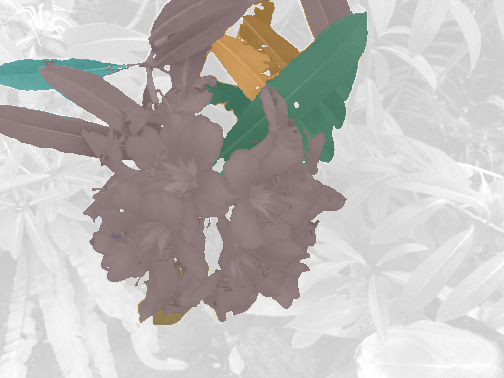}
    \end{subfigure}%
    ~ 
    \begin{subfigure}[t]{0.25\linewidth}
        \centering
        \includegraphics[height=0.8in]{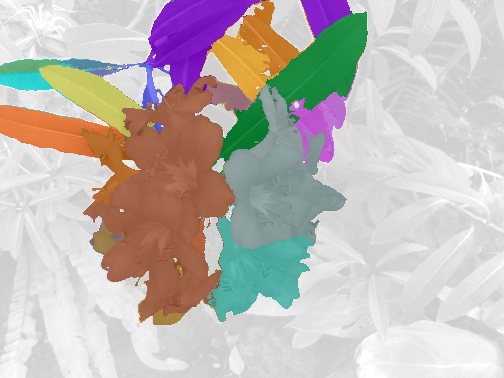}
    \end{subfigure}%
    ~ 
    \begin{subfigure}[t]{0.25\linewidth}
        \centering
        \includegraphics[height=0.8in]{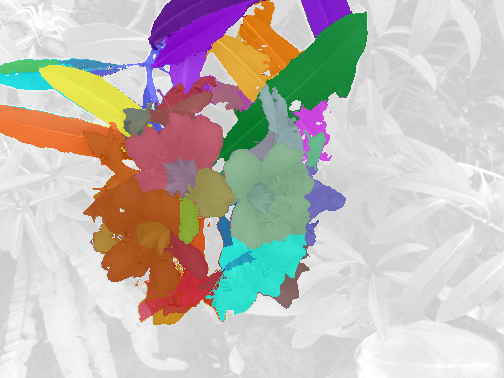}
    \end{subfigure}
    \medskip
    
        \centering
    \begin{subfigure}[t]{0.25\linewidth}
        \centering
        \includegraphics[height=0.8in]{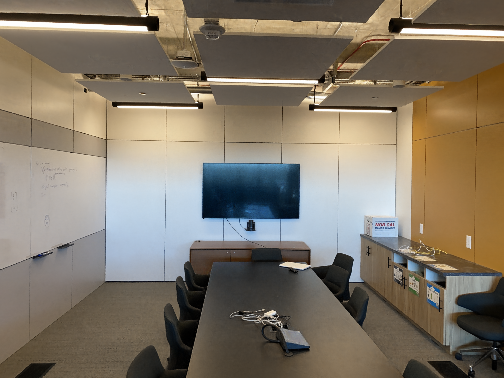}
    \end{subfigure}%
    ~ 
    \begin{subfigure}[t]{0.25\linewidth}
        \centering
        \includegraphics[height=0.8in]{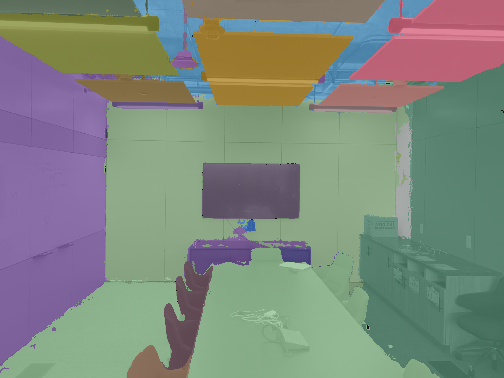}
    \end{subfigure}%
    ~ 
    \begin{subfigure}[t]{0.25\linewidth}
        \centering
        \includegraphics[height=0.8in]{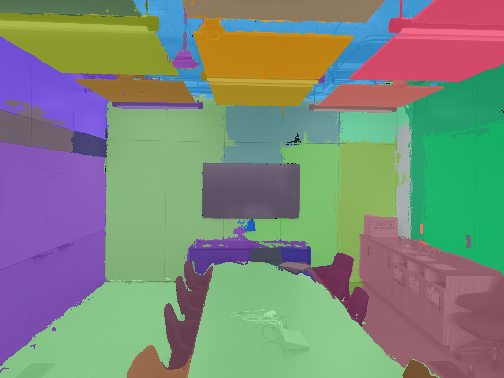}
    \end{subfigure}%
    ~ 
    \begin{subfigure}[t]{0.25\linewidth}
        \centering
        \includegraphics[height=0.8in]{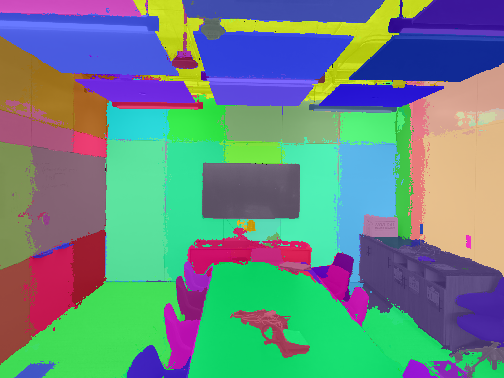}
    \end{subfigure}
    \medskip
    \vspace{-5mm}
    \begin{subfigure}[t]{0.25\linewidth}
        \centering
        \includegraphics[height=0.8in]{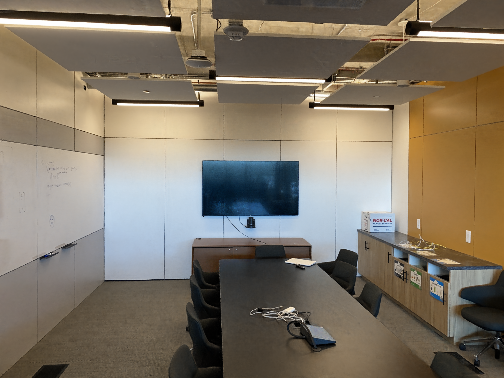}
    \end{subfigure}%
    ~ 
    \begin{subfigure}[t]{0.25\linewidth}
        \centering
        \includegraphics[height=0.8in]{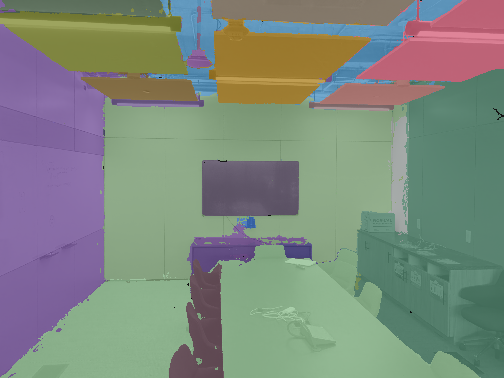}
    \end{subfigure}%
    ~ 
    \begin{subfigure}[t]{0.25\linewidth}
        \centering
        \includegraphics[height=0.8in]{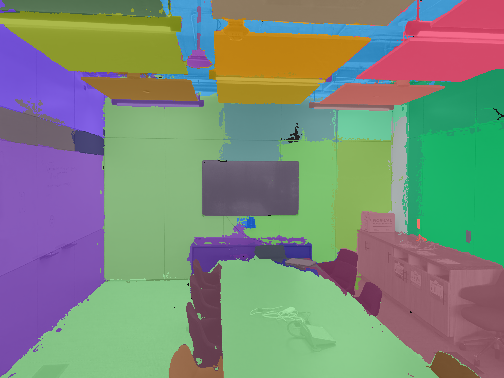}
    \end{subfigure}%
    ~ 
    \begin{subfigure}[t]{0.25\linewidth}
        \centering
        \includegraphics[height=0.8in]{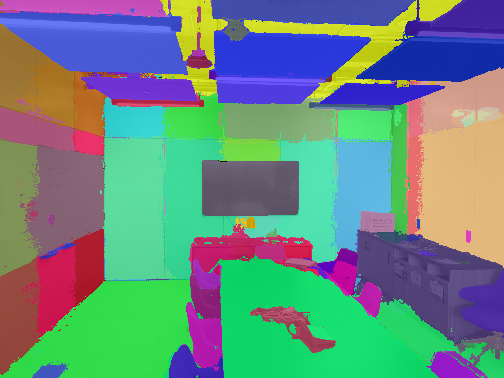}
    \end{subfigure}
    \caption{\textbf{LLFF Dataset}: We showcase the qualitative results on the LLFF dataset.}
    \label{fig:llff}
\end{figure*}

We present qualitative results on the the LLFF dataset~\cite{mildenhall2020nerf} in \cref{fig:llff}.  Our approach is able to generate view-consistent hierarchical segmentation results for real-world scenes. 

% \subsubsection{B.4. ScanNet Dataset}

% We present qualitative results on the the ScanNet dataset in \cref{fig:scannet}. 

\end{document}